%% file: ABSGNN.tex
\documentclass{article}
\usepackage{xcolor}

\usepackage{microtype}
\usepackage{graphicx}
\usepackage{subfigure}
\usepackage{booktabs} 

\usepackage{hyperref}

\input{math_commands.tex}

\usepackage{url}
\usepackage{wrapfig}
\usepackage{multirow}
\usepackage{array}
\newcolumntype{L}[1]{>{\raggedright\let\newline\\\arraybackslash\hspace{0pt}}m{#1}}
\newcolumntype{C}[1]{>{\centering\let\newline\\\arraybackslash\hspace{0pt}}m{#1}}
\newcolumntype{R}[1]{>{\raggedleft\let\newline\\\arraybackslash\hspace{0pt}}m{#1}}
\definecolor{color1}{HTML}{1b9e77}
\definecolor{color2}{HTML}{d95f02}
\definecolor{color3}{HTML}{7570b3}
\definecolor{color4}{HTML}{a6761d}
\definecolor{color5}{HTML}{66a61e}
\definecolor{color6}{HTML}{e6ab02}
\definecolor{color7}{HTML}{386cb0}
\definecolor{color8}{HTML}{e7298a}
\usepackage[accepted]{ABSGNN}

\usepackage{amssymb}
\usepackage{mathtools}
\usepackage{amsthm}

\usepackage[capitalize,noabbrev]{cleveref}

\usepackage{enumitem}
\setlist{nosep}

\theoremstyle{plain}

\theoremstyle{definition}

\theoremstyle{remark}

\usepackage[textsize=tiny]{todonotes}

\newcommand{\revision}[1]{#1}

\newcommand{\cylinder}{\textsc{CylinderFlow}}
\newcommand{\plate}{\textsc{DeformingPlate}}
\newcommand{\airfoil}{\textsc{Airfoil}}
\newcommand{\fonts}{\textsc{InflatingFont}}
\newcommand{\GUN}{\textsc{GraphUNets}}
\newcommand{\flatGNN}{\textsc{MeshGraphNets}}
\newcommand{\linoGNN}{\textsc{MS-GNN-Grid}}
\newcommand{\ours}{\textsc{BSMS-GNN}}

\icmltitlerunning{Efficient Learning of Mesh-Based Physical Simulation with BSMS-GNN}

\begin{document}

\twocolumn[
\icmltitle{Efficient Learning of Mesh-Based Physical Simulation with\\ Bi-Stride Multi-Scale Graph Neural Network}
\icmlsetsymbol{equal}{*}

\icmlsetsymbol{snap-cv}{\dag}
\begin{icmlauthorlist}
\icmlauthor{Yadi Cao}{ucla-cs,snap-cv}
\icmlauthor{Menglei Chai}{google-ar,snap-cv}
\icmlauthor{Minchen Li}{ucla-math}
\icmlauthor{Chenfanfu Jiang}{ucla-math}
\end{icmlauthorlist}

\icmlaffiliation{ucla-cs}{Department of Computer Science, UCLA, Los Angeles, USA}
\icmlaffiliation{google-ar}{AR Perception, Google, Los Angeles, USA}
\icmlaffiliation{ucla-math}{Department of Mathematics, UCLA, Los Angeles, USA}

\icmlcorrespondingauthor{Chenfanfu Jiang}{cffjiangmath.ucla.edu}
\icmlkeywords{Machine Learning, ICML}

\vskip 0.3in
]



\printAffiliationsAndNotice{\textsuperscript{\dag}Work partially done at Snap Inc.}

\input{s_abstract.tex}
\vspace{-10pt}
\input{s_introduction.tex}
\input{s_method.tex}

\input{s_experiment.tex}
\input{s_related_work.tex}
\input{s_conclusion.tex}
\newpage
\input{s_ack}
\bibliography{ABSGNN.bib}
\bibliographystyle{ABSGNN}
\input{s_appendix.tex}

\end{document}

%% file: math_commands.tex
\usepackage{amsmath,amsfonts,bm}

\def\eqref#1{equation~\ref{#1}}

\def\1{\bm{1}}

\def\ve{{\bm{e}}}

\def\vq{{\bm{q}}}

\def\vv{{\bm{v}}}

\def\vx{{\bm{x}}}

\def\mA{{\bm{A}}}

\def\mC{{\bm{C}}}

\def\mG{{\bm{G}}}

\def\mO{{\bm{O}}}

\def\mX{{\bm{X}}}

\DeclareMathAlphabet{\mathsfit}{\encodingdefault}{\sfdefault}{m}{sl}
\SetMathAlphabet{\mathsfit}{bold}{\encodingdefault}{\sfdefault}{bx}{n}

\def\gE{{\mathcal{E}}}

\def\gG{{\mathcal{G}}}

\def\gO{{\mathcal{O}}}

\def\gV{{\mathcal{V}}}



%% file: s_abstract.tex
\begin{abstract}
    Learning the physical simulation on large-scale meshes with flat Graph Neural Networks (GNNs) and stacking Message Passings (MPs) is challenging due to the scaling complexity w.r.t. the number of nodes and over-smoothing.
    There has been growing interest in the community to introduce \textit{multi-scale} structures to GNNs for physical simulation.
    However, current state-of-the-art methods are limited by their reliance on the labor-intensive drawing of coarser meshes or building coarser levels based on spatial proximity, which can introduce wrong edges across geometry boundaries.
    Inspired by the bipartite graph determination, we propose a novel pooling strategy, \textit{bi-stride} to tackle the aforementioned limitations.  Bi-stride pools nodes on every other frontier of the breadth-first search (BFS), without the need for the manual drawing of coarser meshes and avoiding the wrong edges by spatial proximity. Additionally, it enables a one-MP scheme per level and non-parametrized pooling and unpooling by interpolations, resembling U-Nets, which significantly reduces computational costs. Experiments show that the proposed framework, \textit{BSMS-GNN}, significantly outperforms existing methods in terms of both accuracy and computational efficiency in representative physical simulations.
\end{abstract}

%% file: s_introduction.tex
\begin{figure*}[t]
    \centering
    \includegraphics[width = 0.9\linewidth]{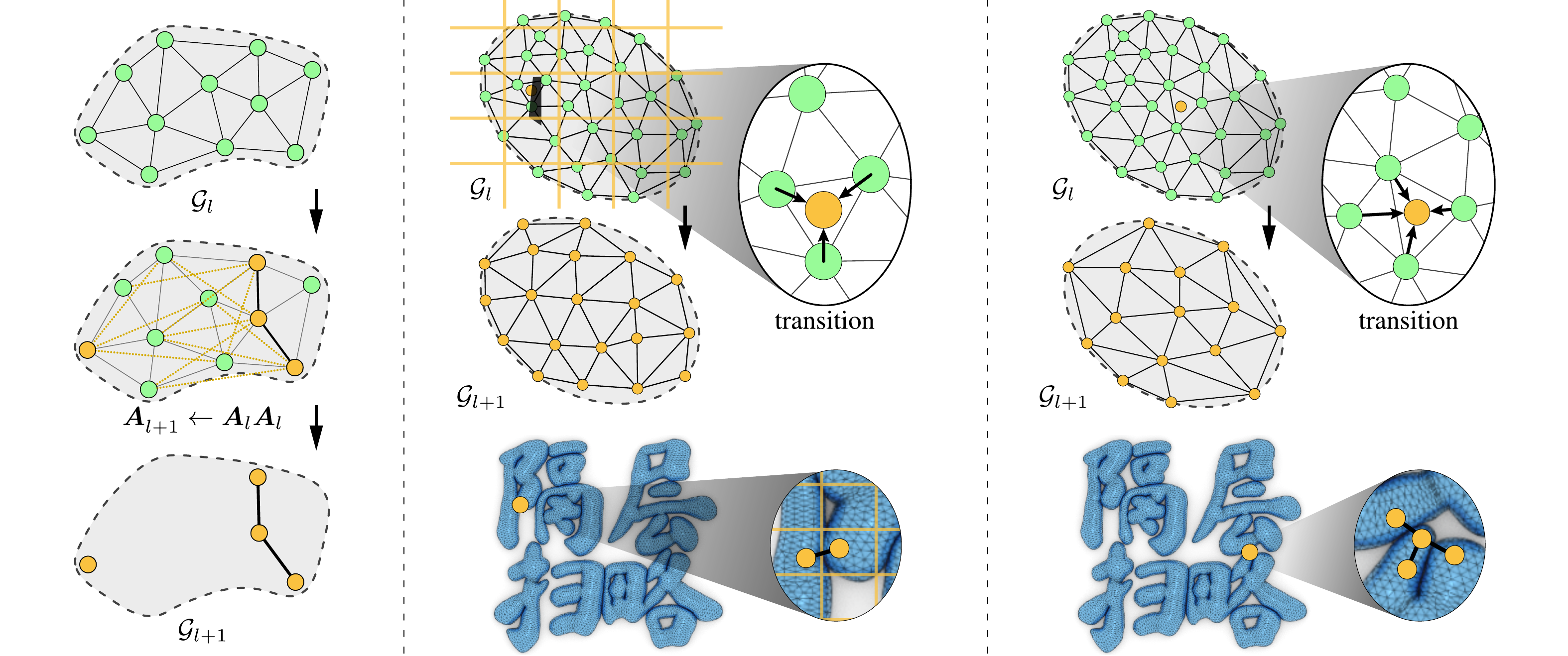}\\
    \setlength{\tabcolsep}{0\linewidth}
    \begin{tabular}{C{0.2191\linewidth}C{0.3154\linewidth}C{0.3154\linewidth}}
        {\footnotesize(a) learnable pooling} & {\footnotesize(b) pooling by rasterization} & {\footnotesize(c) pooling by spatial proximity} \\
    \end{tabular}\vspace{-5pt}
    \caption{\textbf{Issues of existing multi-level GNNs.} (a) A learnable pooling \citep{gao2019graph} may lead to loss of connectivity even after $2^{\text{nd}}$-order enhancement. (b) A pooling by rasterization \citep{lino2021simulating,lino2022towards,lino2022multi} and (c) by spatial proximity \citep{liu2021multi,fortunato2022multiscale} can lead to wrong connections across the boundaries at the coarser level.}\vspace{-10pt}
    \label{fig:compare}
\end{figure*}
\section{Introduction}
\begin{figure*}[t]
    \centering
    \includegraphics[width=0.9\linewidth]{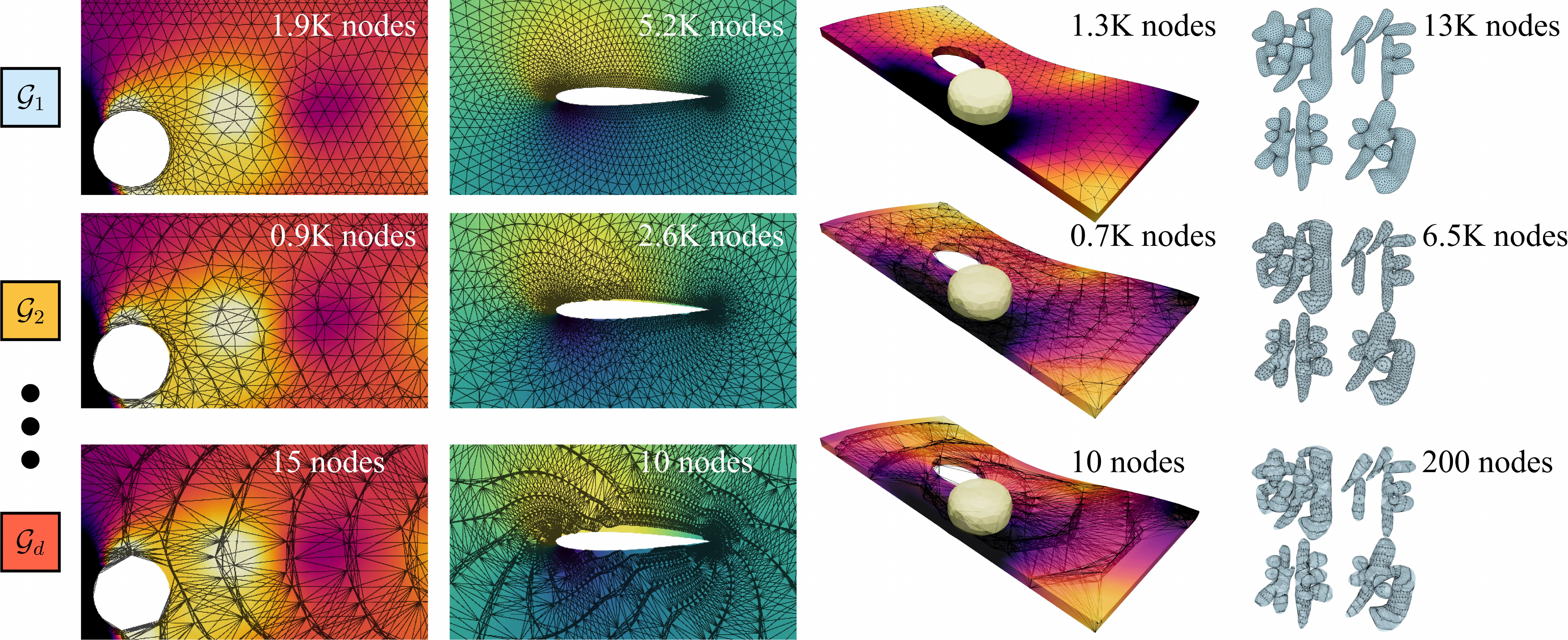}\vspace{2pt}\\
    \setlength{\tabcolsep}{0\linewidth}
    \begin{tabular}{C{0.05\linewidth}C{0.125\linewidth}C{0.275\linewidth}C{0.2\linewidth}C{0.15\linewidth}}
         & {\scriptsize(a) \cylinder~} & {\scriptsize(b) \airfoil~} & {\scriptsize(c) \plate~} & {\scriptsize(d) \fonts~} \\
    \end{tabular}\vspace{-5pt}
    \caption{\textbf{Example multi-level graphs produced by bi-stride pooling.} Our datasets contain both Eulerian and Lagrangian systems. Many meshes are highly irregular and contain massive self-contact, challenging to build coarser-level connections by spatial proximity. Our bi-stride strategy only relies on topological information and is proven to be robust and reliable on arbitrary geometry.}
    \vspace{-10pt}
    \label{fig:example:bistride}
\end{figure*}
Simulating physical systems through numerically solving partial differential equations (PDEs) is a key area in science and engineering, with applications ranging from solid mechanics~\cite{jiang2016material,li2020incremental}, to fluid-~\cite{bridson2015fluid,cao2018liquid}, aerodynamics~\cite{cao2022efficient}, and heat transfer~\cite{cao2019heat}.
However, traditional numerical solvers are often computationally expensive, particularly for time-sensitive applications such as iterative design optimization, where fast online inference is desired.
In recent years, the machine learning community has shown great interest in improving efficiency or replacing traditional solvers with learned models. These works include both end-to-end frameworks~\cite{grzeszczuk1998neuroanimator,obiols2020cfdnet} and those utilizing physics-informed losses~\cite{raissi2019physics,karniadakis2021physics,sun2020surrogate}.
Many existing works apply convolutional neural networks (CNNs)~\cite{fukushima1982neocognitron} to physical systems residing on two- or three-dimensional structured grids~\cite{guo2016convolutional,tompson2017accelerating,kim2019deep,fotiadis2020comparing}. However, the strict dependency on regular domain shapes makes it non-trivial to be applied on unstructured meshes. While it is possible to deform simple irregular domains into rectangular shapes to apply CNNs~\cite{gao2021phygeonet,li2022fourier}, the challenge remains for domains with complex topologies, which are common in practice.

As a result, the use of graph neural networks (GNNs) in physics-based simulations on unstructured meshes has gained significant attention in recent years~\cite{battaglia2018relational,sanchez2018graph,belbute2020combining,pfaff2020learning,sanchez2020learning,harsch2021direct,gao2022physics}.
The naive GNN approach stacks multiple MPs to model information propagation throughout space. However, as the graph size increases, this approach faces two major challenges:
(1) \textbf{Complexity}: as both the number of nodes to be processed and the MP iterations increase linearly, a quadratic complexity becomes inevitable for both the running time and memory usage of the computational graph~\cite{fortunato2022multiscale}.
(2) \textbf{Oversmoothing}: the graph convolution can be seen as a low-pass filter that suppresses the higher-frequency signals~\cite{chen2020measuring, li2020multipole}. Then the stacked MPs iteratively project the information onto the eigenspace of the graph while all higher-frequency signals are smoothed out, making the training more difficult.

To address these limitations, researchers have begun introducing multi-scale GNNs (MS-GNNs) for physics-based simulation~\cite{li2020multipole,liu2021multi,lino2021simulating,fortunato2022multiscale,lino2022multi,lino2022towards}. The multi-scale approach mitigates over-smoothing by building sub-level graphs at coarser resolutions, resulting in longer-range interactions and fewer MP iterations.
The existing approaches for constructing the multi-scale structure include: utilizing spatial proximity to generate sub-level graphs at coarser levels~\cite{lino2021simulating,liu2021multi,lino2022towards}; applying Guillard's coarsening algorithm~\cite{guillard1993node}~\cite{lino2022multi}; manually drawing coarser meshes for the original geometry~\cite{liu2021multi,fortunato2022multiscale}; or randomly pooling nodes and applying factorization to the adjacency matrix~\cite{li2020multipole}. However, these solutions all have their limitations.
For example, learnable or random pooling can introduce artificial partitions in the sub-level graphs (Fig.~\ref{fig:compare}. (a)), even with adjacency enhancement, which impedes information exchange across partitions, spatial proximity can lead to wrong edges across the boundaries at coarser levels (Fig.~\ref{fig:compare}. (b) and (c)); Guillard's algorithm only applies to 2D triangle meshes; and manually drawing tens of thousands of meshes is too labor-intensive.
We observe that all these limitations originate from the unmatured operations: \textit{pooling} and \textit{building graph connections at coarser levels}. We desire to design operations that: 1) conserve the correct connections at coarser levels, 2) do not introduce edges that blur the boundaries, 3) are general for any mesh type, and 4) are automatic.

We tackle these challenges with two progressive contributions:
\begin{itemize}
    \item \textbf{First}, we introduce a novel yet simple pooling strategy, \textit{bi-stride}. Bi-stride is inspired by the bi-partition determination in DAG (directed acyclic graph). It pools all nodes on every other BFS (breadth-first-search) frontier, such that a $2^{\text{nd}}$-powered adjacency enhancement \revision{($\mA\leftarrow\mA^2$, where $\mA$ is the adjacency matrix of the graph)} conserves all the correct connectivity. Bi-stride solely uses the input mesh without the need for spatial proximity, is general for any mesh type, and is fully automatic.
    \item \textbf{Second}, bi-stride pooling conserves direct connections between pooled/un-pooled nodes; Utilizing this advantage, one MP suffices to exchange the information between pooled and unpooled nodes before moving into the adjacent level; We also design a non-parameterized aggregating and returning method, resembling the interpolation in \revision{U-Net}, to handle the transition between adjacent levels. These simplifications significantly reduce the computational requirements compared to SOTAs.
\end{itemize}
Together, these two contributions give birth to our \textit{Bi-Stride Multi-Scale GNN} (\textit{BSMS-GNN}), a novel framework representing a significant advancement in the field of learned mesh-based simulations, particularly for the deployment in real industrial applications where meshes are often complex in geometry and large in size.

%% file: s_method.tex
\section{Multi-Scale Building as Preprocessing}
\label{sec:preproc}
\input{ss_method_bistride.tex}
\section{Bi-Stride Multi-Scale (BSMS)-GNN}
\input{ss_method_define.tex}
\input{ss_method_transit.tex}

%% file: ss_method_bistride.tex
We first introduce the bi-stride pooling strategy, multi-scale building, and the corresponding data preprocessing. Note that all algorithms and preprocessing steps in this section are deterministic and done in one pass.
\subsection{Motivations}
As summarized in Fig.~\ref{fig:compare}. The pooling strategy can be categorized into two groups by either utilizing spatial proximity~\cite{lino2021simulating,lino2022towards} or purely relying on graph information~\cite{gao2019graph,li2020multipole}. For complex geometry, it's preferred to avoid spatial proximity unless desired by the simulation cases (such as contact or interface interactions). The only benchmark that uses only graph information is \GUN~\cite{gao2019graph} with an obvious drawback: it's easy to lose the connectivity and artificially introduce partitions, even with adjacent matrix enhancement (Fig.~\ref{fig:compare} (a)).

For a more clear illustration, we first define the adjacency enhancement by the $K^{\text{th}}$-order matrix power as $\mA\leftarrow\mA^K$, where $\mA$ is the adjacency matrix of the graph. Geometrically, $\mA(i,j)=1$ means the edge $(i,j)$ exists, and $\mA^K(i,j)=1$ means that node $j$ is connected to node $i$ via at most $K$ hops. Given a pooling strategy $\text{P}$ and \revision{the pooled nodes' indices $\mathcal{I}$}, we define a $K{\text{th}}$-order outlier set as $\gO_K$, where the nodes in $\gO_K$ are not connected to any pooled nodes even after $K^{\text{th}}$-order adjacency enhancement: $\mA^K(i,j)=0,\forall i\in \revision{\mathcal{I}},\forall j\in\gO_K$.

We finally define that a pooling strategy $\text{P}$ is $K^{\text{th}}$-order connection conservative (K-CC) if $\gO_K$ is empty. Empirically, the larger $K$ in $K^{\text{th}}$-order adjacency enhancement is harmful to distinguish the node features: as $K$ increases, $\mA^K(i,j)$ (converted to boolean) approaches a matrix with all its entries equal to $1$, representing a fully connected graph; then a single convolution will average all node features and make them indistinguishable. The most preferred, i.e. the smallest possible $K$ we seek is naturally $2$.
With such reason, \citet{gao2019graph} uses the smallest $2^{\text{nd}}$ order enhancement to help conserve the connectivity. Nonetheless, there is no theoretical guarantee that a learnable pooling module is consistently 2-CC for any graph. These limitations motivate us to create a consistent 2-CC pooling strategy, as followed in Sec.~\ref{sec:bistride:adjenhance}.

\subsection{Bi-Stride Pooling and Adjacency Enhancement}
\label{sec:bistride:adjenhance}
\begin{figure}[t]
    \centering
    \includegraphics[width = 0.8\linewidth]{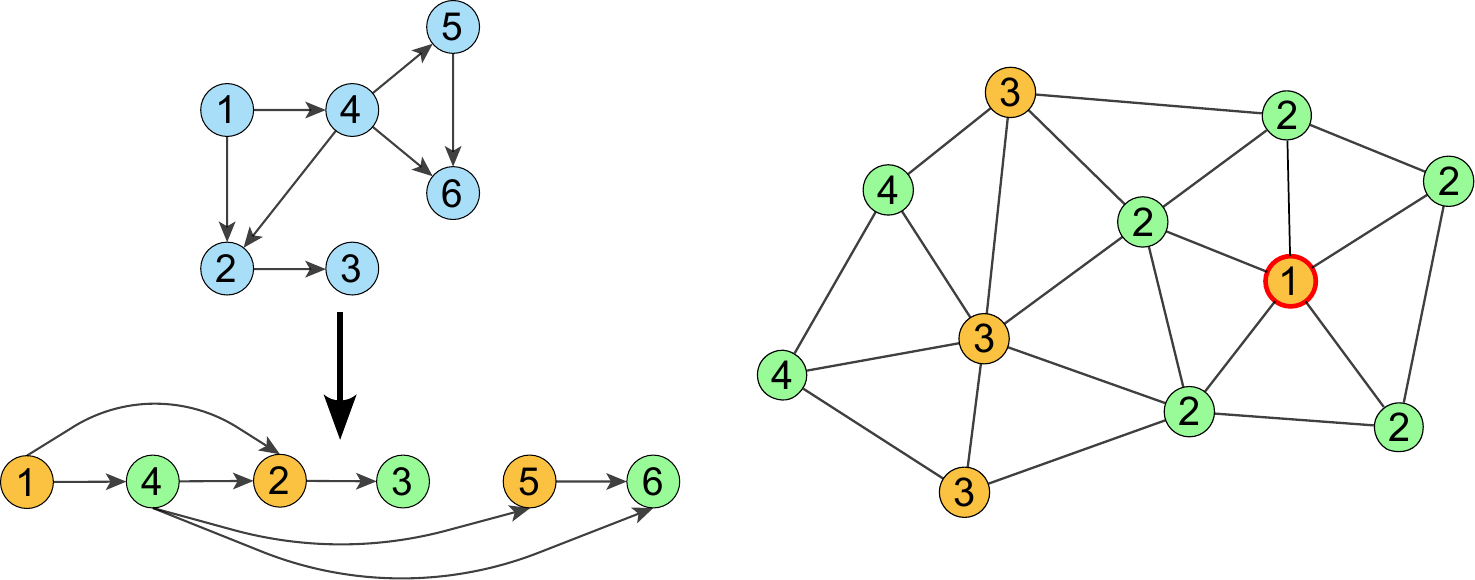}\vspace{2pt}\\
    \setlength{\tabcolsep}{0\linewidth}
    \begin{tabular}{C{0.04\linewidth}C{0.43\linewidth}C{0.43\linewidth}}
         & {\scriptsize(a) Bi-partition of a DAG} & {\scriptsize(b) Bi-stride of a mesh} \\
    \end{tabular}\vspace{-5pt}
    \caption{\textbf{The similarity between Bi-partition and Bi-stride.} The number in the circle means the depth of the frontiers of either topological sorting or BFS. The red-bounded circle (number 1) means the starting node, i.e. the seed.}
    \vspace{-10pt}
    \label{fig:DAG}
\end{figure}
We draw the initial inspiration from the bi-partition determination algorithm~\cite{asratian1998bipartite} in a directed acyclic graph (DAG). As shown in Fig.~\ref{fig:DAG}(a), after topological sorting, pooling on every other depth (yellow and green) generates a bi-partition where all edges reside between two partitions, and pooling either partition is obviously 2-CC.
To resemble bi-partition determination on a mesh, which is not bi-partite due to cycles, we can conduct a breadth-first search (BFS) to compute the geodesic distances from a seed to all other nodes, and then stride and pool all nodes at every other BFS frontier (\textit{bi-stride}). A bi-stride example is shown in Fig.~\ref{fig:DAG}(b), where the number in each vertex represents the geodesic distance to the seed (node $1$ in the red circle) by BFS. This pooling is 2-CC by construction and conserves direct connections between pooled and unpooled nodes. As a result, we avoid building edges by using spatial proximity or handling cumbersome corner cases such as cross-boundary connections while also using the smallest adjacency enhancement.

\paragraph*{Seeding Heuristics}
We claim that there should exist some freedom as long as the seeding is balanced to a certain degree. For training datasets, we choose two deterministic seeding heuristics: 1) closest to the center of a cluster (CloseCenter) for \fonts, and 2) the minimum average distance (MinAve) for all other cases, and we preprocess the multi-level building in one pass. One can consider the cheaper heuristic CloseCenter during the online inferring phase if an unseen geometry is encountered. The details of the algorithms can be found in Sec.\ref{sec:appdx:algo}. \revision{The sensitivity study on choosing a different seeding heuristic in the test phase is presented in Sec.\ref{sec:appdx:abaltion:seeding}.}

\paragraph*{Contact Edges}
For problems involving contacts, such as \plate~(Fig.~\ref{fig:example:bistride}(c)) and \fonts~(Fig.~\ref{fig:example:bistride}(d)), the finest-level contact edges $\mA^C$ are built dynamically by spatial proximity between nodes. Note the edge-building by the spatial proximity does not apply to the internal elastic mechanics, whose edge is defined by the mesh only. The enhancement of the contact edges should be handled properly for multi-scale GNN, which, to the best of our knowledge, has not been addressed in prior works. At any level $l$, given adjacent matrices $\mA_l$ obtained by the input mesh and the enhancement rule, and the contact edge $\mA^C_l$ at this level, we first apply bi-stride pooling to select nodes $\mathcal{I}$, then enhance $\mA_{l+1}$ and $\mA^C_{l+1}$ using the following rule, \revision{where $[\mathcal{I},\mathcal{I}]$ means striding on the matrix rows and columns}:
\begin{equation}
    \label{eq:adj:enhance}
    \begin{split}
        \mA'_{l+1} \leftarrow \mA_l \mA_l, \quad \mA_{l+1} & \leftarrow \mA'_{l+1}[\mathcal{I},\mathcal{I}],\\
        \mA'^C_{l+1} \leftarrow \mA_l \mA^C_l \mA_l , \quad \mA^C_{l+1} & \leftarrow \mA'^C_{l+1}[\mathcal{I},\mathcal{I}].
    \end{split}
\end{equation}
The enhancement of contact edges can be geometrically interpreted as contact edge $(i,j)$ should exist if $j$ is reachable from $i$ in 2 hops and at least one of them is a contact edge at the finer level. We prove in Sec.~\ref{sec:appdx:prove:contact:conserve} that bi-stride pooling with the enhancement in Eq.~\ref{eq:adj:enhance} also conserves all contact edges.

%% file: ss_method_define.tex
\begin{figure*}[t]
    \centering
    \includegraphics[width = 0.9\linewidth]{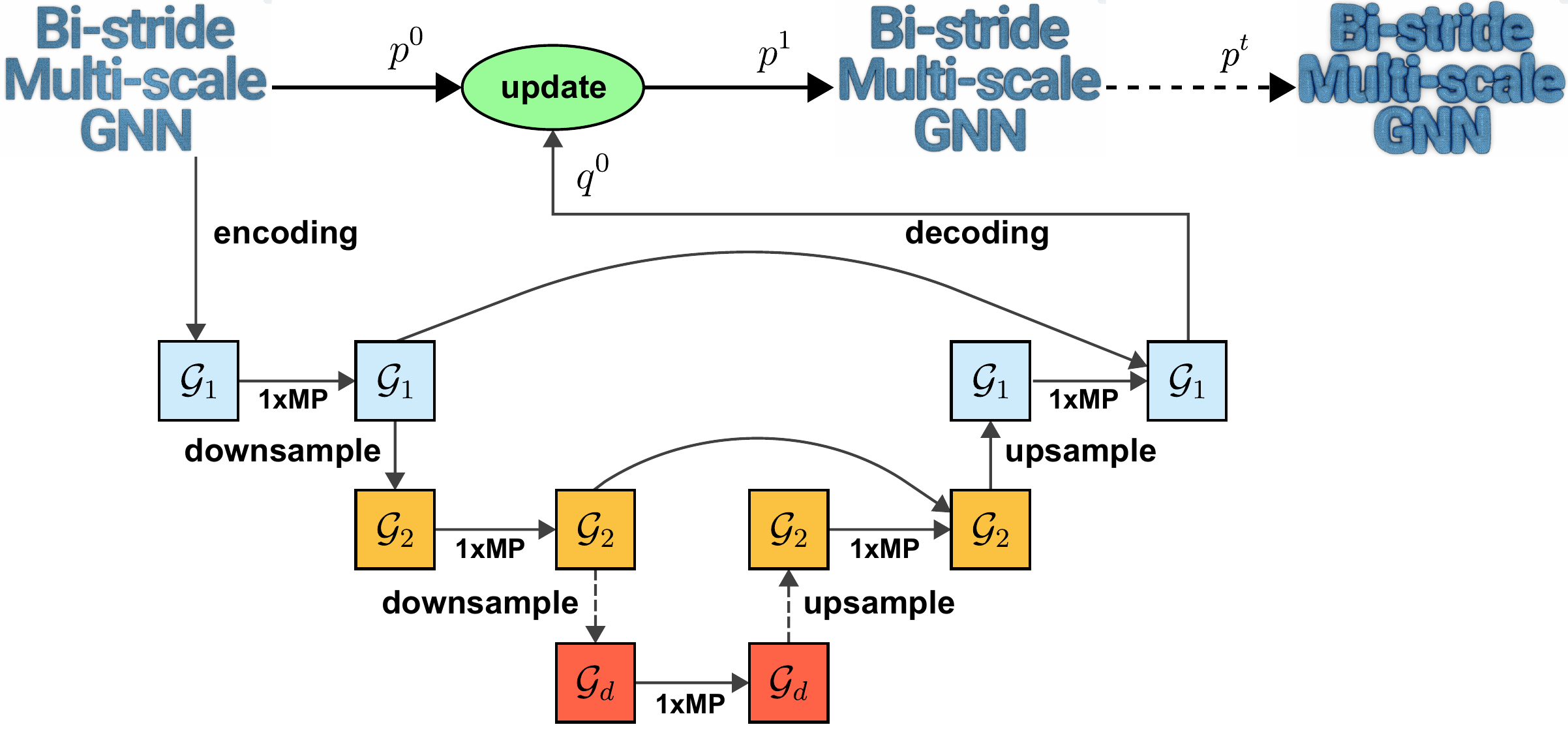}\vspace{-5pt}
    \caption{\textbf{\ours~pipeline} is trained with one-step supervision. $\gG_1,\gG_2,\cdots,\gG_d$ are graphs at different levels (fine to coarse). Encoder/decoder only connects the input/output fields with the latent fields at $\gG_1$. Latent nodal fields are updated by one MP at each level. The bi-stride pooling selects the pooled nodes for the adjacent coarser level, and the transition is conducted in a non-parameterized way.}\vspace{-10pt}
    \label{fig:scheme}
\end{figure*}

Here we formally introduce \ours, a hierarchical GNN where the multi-level structure has been determined by the input mesh and the preprocessing in Sec.~\ref{sec:preproc}.
\subsection{Definitions}
\label{sec:define}
Figure.~\ref{fig:scheme} presents the overall structure of \ours. We consider the evolution of a physics-based system discretized on a mesh, which is converted to an bi-directed graph $\gG_1=\left(\gV_1,\gE_1\right)$. Here, with subscript $1$, $\gV_1$, and $\gE_1$ label the nodal fields and the edges at the finest level (the input mesh), respectively. Specifically for edges, we define $\gE_1=\{\gE_1^1,\cdots,\gE_1^S\}$, where $\gE_1^1$ is the edge set directly copied from the input mesh and $\{\gE_1^k\vert_{k=2}^S\}$ are the optional problem-dependent edge sets. For example, both \plate~(Fig.~\ref{fig:example:bistride}(c)) and \fonts~(Fig.~\ref{fig:example:bistride}(d)) benchmarks have a contact edge set $\gE_1^2$ for the colliding vertices. We use $\{p,q\}$, stacked vectors of $\{p_i,q_i\}$ of all nodes $i\in\gV_1$, to denote the input and output nodal fields, respectively. Given an input field $p^{j}$ at a time $t_j$, one pass of \ours~returns the output field $q^{j+1}$ at time $t_{j+1}=t_j+\Delta t$, where $\Delta t$ is a fixed time step size. The output $q$ can contain more physical fields than the input $p$ and must be able to derive the input for the next pass. The rollout refers to iteratively conducting \ours~from the initial state $p^0 \to q^1 \to p^1 \to \cdots \to q^n$ and producing the temporal sequence output $\{q^1,q^2,\cdots,q^n\}$ within the time range of $\left(t_0,t_0+n\Delta t\right]$, where $n$ is the total number of evaluations.

\paragraph*{Message Passing} In general, we follow the \textit{encode-process-decode} fashion in \flatGNN, where encoder and decoder only appear at the top level $\gG_1$, mapping the nodal input $p$ and output $q$ to/from the latent feature $\vv$, respectively (see Table~\ref{tab:model:inputs} for the domain-specific information). As for the processor, unlike~\cite{fortunato2022multiscale}, we observe that a single MP per level is sufficient for all experiments. We do not separately encode the edge offsets $\Delta\vx_{ij}=\vx_i-\vx_j$, instead, simply prepend this to the stacked sender/receiver latent as the input to calculate edge flow. For a problem involving $S$ edge sets, an MP pass at level $l$ is formulated as:
\begin{equation}
    \label{eq:MP}
    \begin{split}
        \ve_{l,ij}^{s}&\leftarrow\text{f}^{s}_l\big(\Delta\vx_{l,ij},\vv_{l,i},\vv_{l,j}\big),\quad s=1,\cdots,S,\\
        \vv'_{l,i}&\leftarrow\vv_{l,i}+\text{f}^{V}_l\Big(\vv_{l,i},\sum_{j}\ve_{l,ij}^{1},\cdots,\sum_{j}\ve_{l,ij}^{S}\Big),
    \end{split}
\end{equation}
where $\text{f}$ is a MLP function, $\ve$ is the latent information flow through an edge, and $\vv$ is the latent node feature. Please refer to Sec.~\ref{sec:appdx:arch} for the detailed architecture of the model.

%% file: ss_method_transit.tex
\subsection{Transition Between Levels}
\label{sec:transition}
We handle information transition between two adjacent levels with non-parametrized \textit{downsampling} and \textit{upsampling} modules to reduce the overhead of the learnable transition modules between every pair of adjacent levels. Here we define downsampling as the sequence of pooling nodes and then aggregating the information from the neighbors to the coarser level, and upsampling as the sequence of unpooling and then returning the information of the pooled nodes to their neighbors at the finer level.
\begin{figure}[ht]
    \centering
    \includegraphics[width = 0.75 \linewidth]{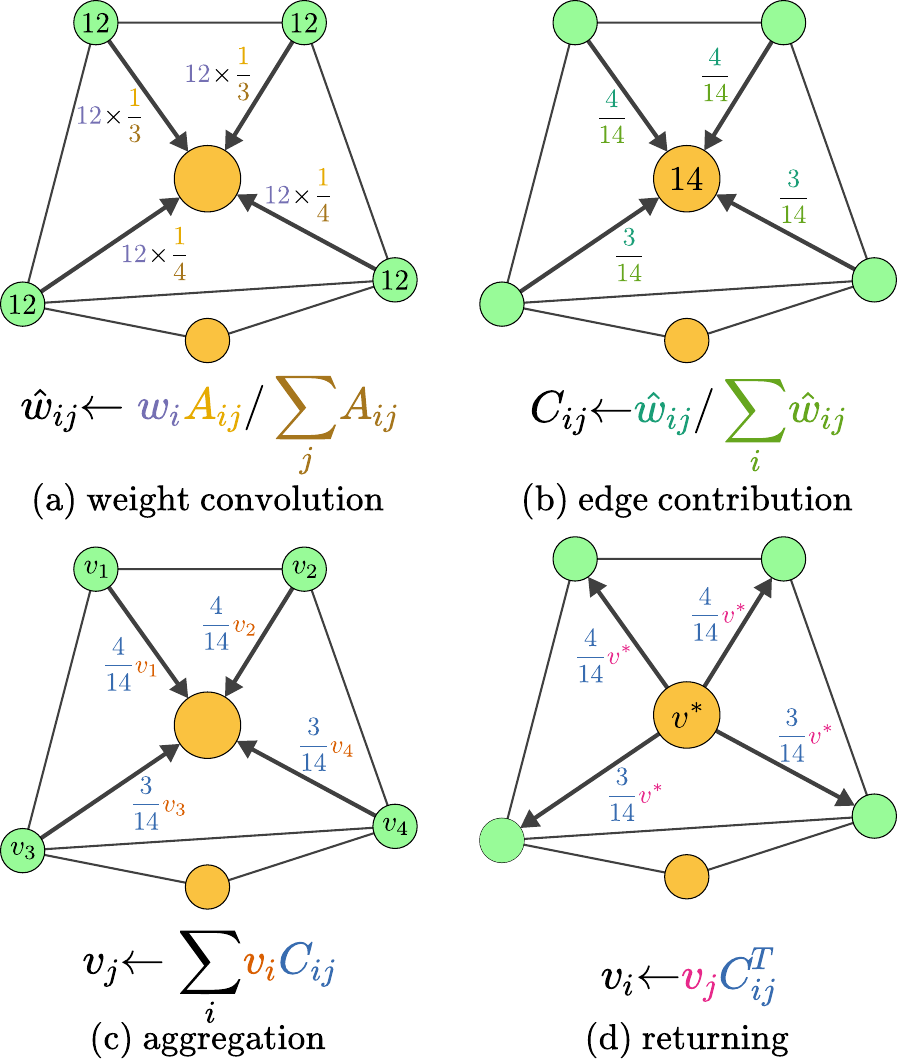}\\
    \caption{\textbf{Schematic plot of transition between adjacent levels.}}\vspace{-10pt}
    \label{fig:transit}
\end{figure}

\paragraph*{Downsampling} Let $\mA$ be the unweighted, boolean adjacency matrix. We initiate a nodal weight field $w$, which is one on the finest level and will be aggregated during downsampling. With the receiver $j$ and its sender $i$, the formal procedure of downsampling is formulated as (Fig.~\ref{fig:transit}):
\begin{itemize}
    \item Normalize by row as in a standard graph convolution $\hat{\mA}_{ij}\leftarrow\mA_{ij}/\sum_j\mA_{ij}$, and then convolve the weight once $\hat{w}_{ij}\leftarrow w_i\hat{\mA}_{ij}$ (Fig.~\ref{fig:transit}(a));
    \item Calculate edge weights $\mC_{ij}\leftarrow\hat{w}_{ij}/\sum_i\hat{w}_{ij}$, where $\mC$ can be viewed as a contribution table with $\mC_{ij}$ as the share of weights in receiver $j$ contributed by sender $i$ (Fig.~\ref{fig:transit}(b));
    \item Convolve the latent information by contribution table $\vv_j\leftarrow\sum_i\vv_i\mC_{ij}$, which is equivalent to equally splitting and sending the weighted information out from senders, and then conducting the weighted average on receivers (Fig.~\ref{fig:transit}(c)).
\end{itemize}
\paragraph*{Upsampling} After unpooling, all nodes except pooled ones have zero information. A returning process, resembling transposed convolution \revision{in U-Net}, can help distinguish the information between receivers. With the contribution table $\mC$ recording edge weights, a natural choice is $\vv_i\leftarrow\vv_j\mC^T_{ij}$ (Fig.~\ref{fig:transit}(d)).

%% file: s_experiment.tex
\section{Experiments}
\label{sec:experiments}

\urlstyle{same}

\subsection{Experiment Setup}
\label{sec:experiments:setup}
\paragraph*{Datasets} We adopt three representative public datasets from GraphMeshNets~\cite{pfaff2020learning}: 1) \cylinder: incompressible fluid around a cylinder; 2) \airfoil: compressible flow around an airfoil; and 3) \plate: deforming an elastic plate with an actuator. In addition, we create a new dataset, \fonts, featuring the inflation of enclosed elastic surfaces~\cite{fang2021guaranteed}. The example plots of them are plotted in Fig.~\ref{fig:example:bistride}.
Compared to existing datasets, \fonts~has more complex geometric shapes, 2 to 8 times the number of nodes, and 70 times the number of contact edges. Hence it is very suitable for testing the capability of competing GNNs in terms of scalability and compatibility with complex geometries. More details are included in Sec.~\ref{sec:appdx:datasets}.
\paragraph*{Baselines} On all datasets, we compare the computational complexity, training/inference time, and memory footprint of \ours~to baselines: 1) \flatGNN~\cite{pfaff2020learning}: the single-level GNN architecture of GraphMeshNets; 2) \linoGNN~\cite{lino2021simulating,lino2022towards,lino2022multi}: a representative work for those building the hierarchy with spatial proximity; and 3) \GUN~\cite{gao2019graph}: a representative work for those using learnable modules for pooling. The reimplementation details can be found in Sec.~\ref{sec:appdx:arch}. We note again that methods such as \cite{liu2021multi,fortunato2022multiscale} are not practical because they require manually drawing coarser meshes for every trajectory instance with CAE software. For all cases combined, this means manually drawing about $20,000$ meshes.
\paragraph*{Implementation} We implement our framework with PyTorch~\cite{pytorch19lib} and PyG (PyTorch Geometric)~\cite{pygeo19lib}. We train the entire model by supervising the single-step $L_2$ loss between the ground truth and the nodal field output of the decoding module. More details, such as the network structures and hyperparameters are included in~\ref{sec:appdx:arch}.
Our datasets and code are publicly available at \textit{\url{https://github.com/Eydcao/BSMS-GNN}}.
\paragraph*{MISCs} The ablation study is conducted for the specific choice of our transition method in Sec.~\ref{sec:appdx:abaltion}. The ablation study concerning whether or not to use learnable pooling modules is not standalone listed but covered by comparing to \GUN~in full-scale experiments (details in Sec.~\ref{sec:experiments:results}).

\subsection{Results and Discussions}
\label{sec:experiments:results}
By evaluating \ours~and other competitors on all the baselines (Sec.~\ref{sec:experiments:setup}), we observe the following conclusions:
\begin{itemize}
    \item We experimentally show the learnable pooling method is not applicable for the deployment of large-scale, complex geometries;
    \item We design a small-scale experiment where pooling by spatial proximity leads to wrong edge and inference;
    \item \ours~shows dominant advantages in significantly less memory footprint, training time to reach the desired accuracy, and the inference time;
    \item \ours~also reaches the highest accuracy, reducing the rollout RMSE approximately by half on \fonts~with the largest mesh size and the most complex geometries; we also zero-shoot the trained model on a teaser with approximately 7x more nodes with the same level of accuracy.
\end{itemize}
For conciseness, we plot the detailed results in Table.~\ref*{tab:detail:res} and Table.~\ref*{tab:detail:res2} in Sec.~\ref{sec:appdx:detail:res}.
\begin{figure}[t]
    \centering
    \includegraphics[width = \linewidth]{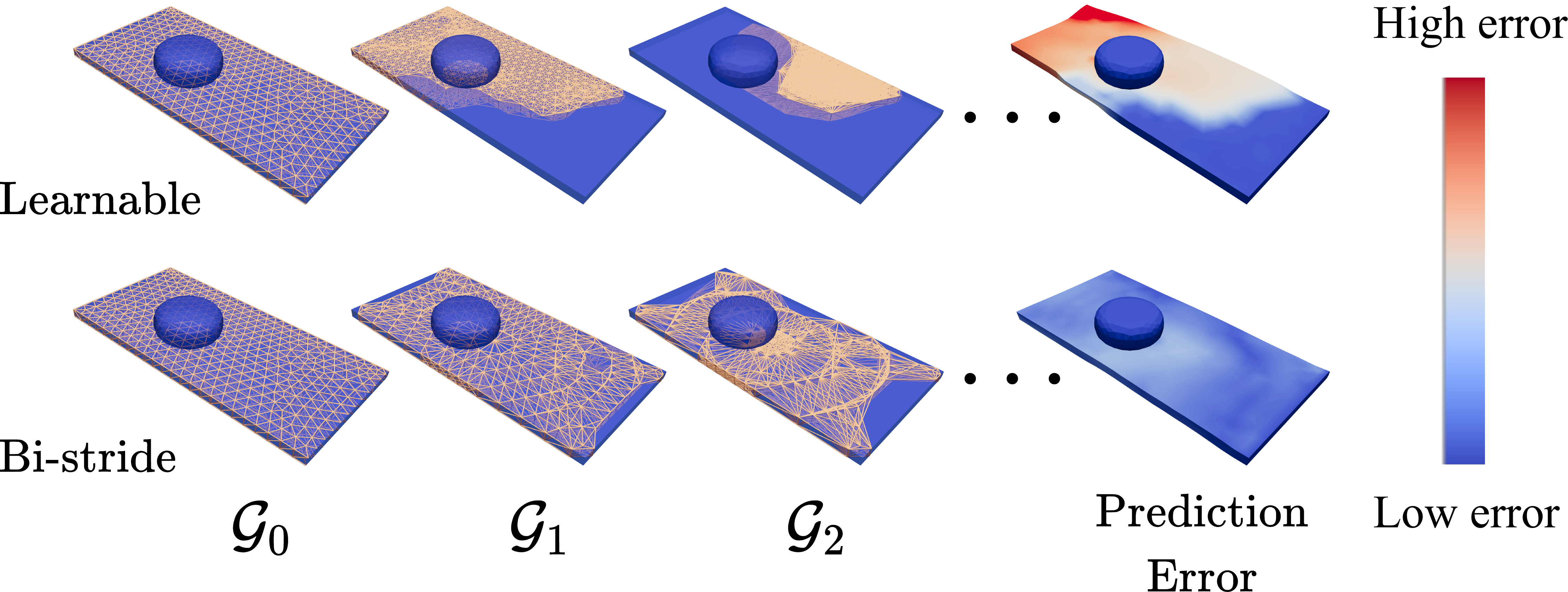}\vspace{-10pt}
    \caption{\textbf{\ours~Comparison between pooling methods}. Unlike bi-stride pooling, which generalizes to any input graph, a learnable module leads to unfair pooling on unseen geometries, impeding the information exchange for unselected nodes. The inferred results show larger errors than bi-stride pooling.}\vspace{-10pt}
    \label{fig:compare:pool}
\end{figure}

\paragraph*{Disadvantages of Learnable Pooling} Compared to other competitors, \GUN~shows a significantly higher RMSE in both 1-step and global rollouts on all datasets, except for the \airfoil~dataset, where the mesh is \revision{consistent} across trajectories. To confirm varying mesh leads to poor inference with learnable pooling, we apply the trained \GUN~model to an unseen test trajectory of the \plate~dataset.  Fig~\ref{fig:compare:pool} clearly reveals the unfair pooling distribution by the learnable module, which impedes information passing on coarser levels and results in poor inference. In comparison, bi-stride generates uniform pooling and accurate inference.

Additionally, \GUN~has to conduct the adjacency matrix multiplication in the forward pass, which results in a 2-40x increase in the training and inference times, particularly in larger datasets. In the largest \fonts, one training epoch requires nearly 50 hours to complete, making the convergence of the model infeasible. Given its poor performance in both accuracy and efficiency, we conclude that \GUN~is not suitable for simulation cases with large-scale, complex geometries. By default, we will not specifically make comparisons to \GUN~in the following discussions.

\begin{figure*}[t]
    \centering
    \includegraphics[width = 0.9\linewidth]{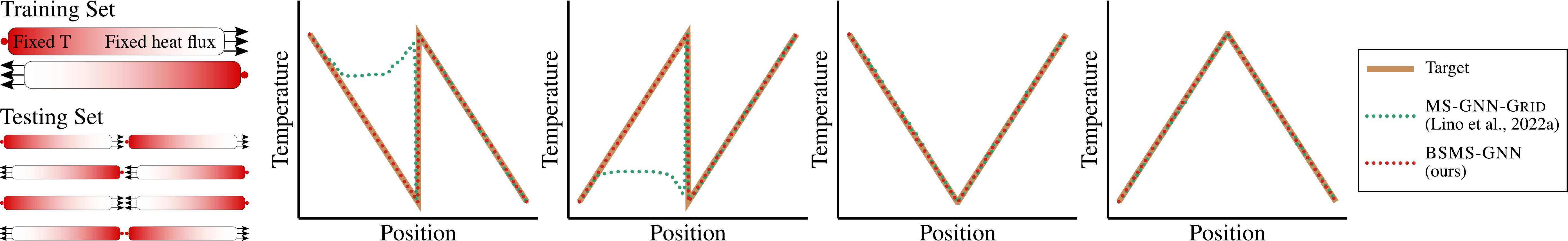}\vspace{-5pt}
    \caption{\textbf{Failure cases for \linoGNN.} Left: the configuration of the simplest failure case for multi-level GNNs by spatial proximity: steady-state 1-D heat transfer. Right, leading two columns: two tests showing that even if trained to convergence, the erroneous edge across the boundary can still result in the wrong inference. Right, last two columns: the erroneous edge coincidentally does not affect the results due to the symmetry of the solution, and no heat will diffuse between two nodes with the same temperature.}\vspace{-10pt}
    \label{fig:failure:1d}
\end{figure*}

\begin{figure}[t]
    \centering
    \includegraphics[width = 0.9\linewidth]{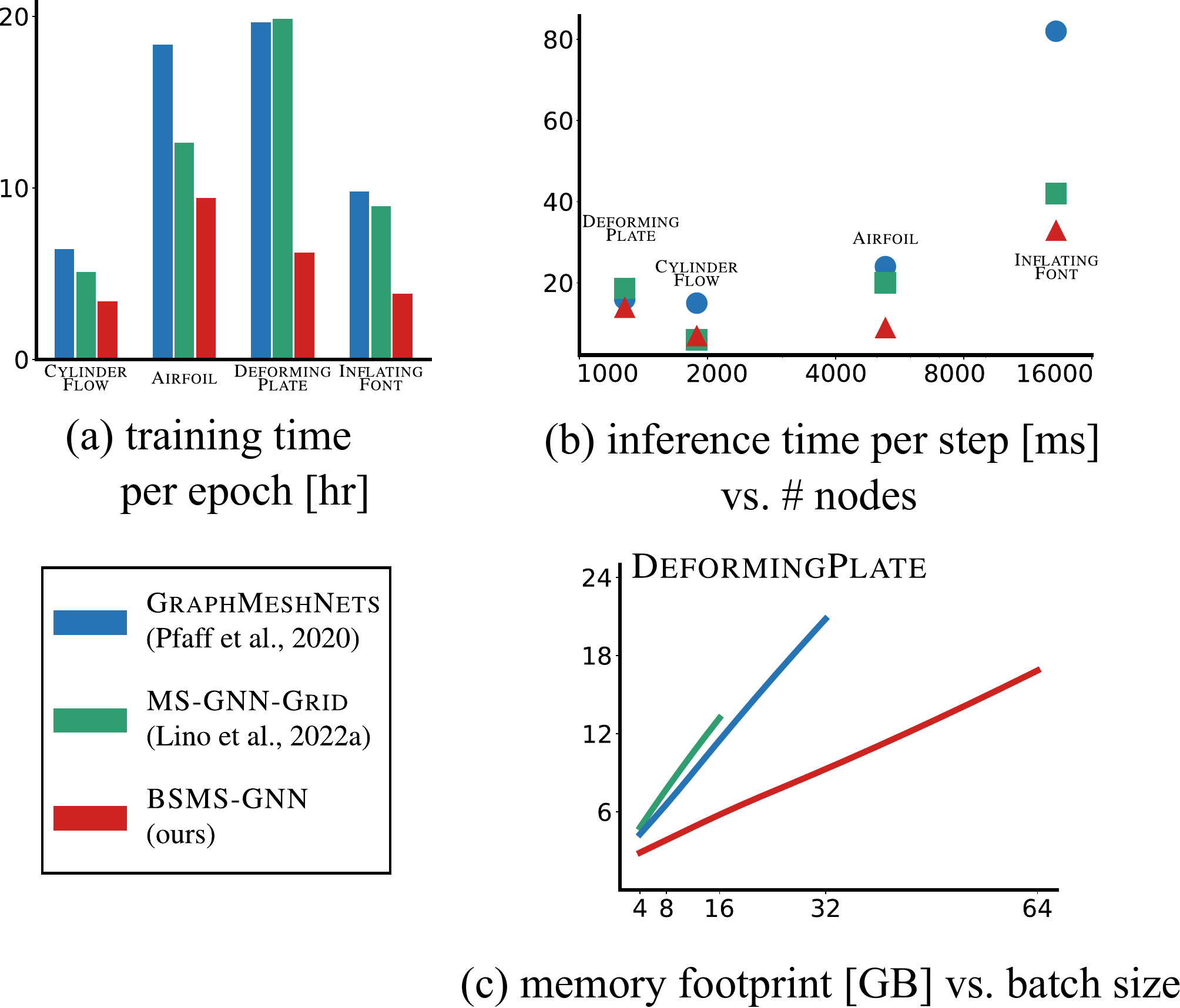}\vspace{-5pt}
    \caption{\textbf{Performance comparison between \ours, \linoGNN, and \flatGNN.} \GUN~is not included because of poor performance. Full results are plotted in Table.~\ref*{tab:detail:res} and Table.~\ref*{tab:detail:res2} in Sec.~\ref{sec:appdx:detail:res}.}\vspace{-10pt}
    \label{fig:perform:compare}
\end{figure}

\paragraph*{Failure Cases for Spatial Proximity} To illustrate the adversarial impact of wrongly constructed edges by spatial proximity, we design a simple 1-D steady-state heat transfer simulation on sticks (Figure~\ref{fig:failure:1d} left), on which \ours~and \linoGNN~are trained and evaluated. The training set consists of two mirrored instances, where one end of the stick is fixed at a certain temperature and the other end has a fixed heat flux, resulting in a linear temperature distribution. In the test set, we simply align two sticks in a head-to-tail configuration with a tiny space between them to prevent heat diffusion across the boundary. \linoGNN, utilizing spatial proximity, incorrectly constructs an edge between the two sticks. As a result, in half of the test cases, \linoGNN~shows wrong results at the boundary due to the erroneous edge(Figure~\ref{fig:failure:1d} right, leading two columns). Although only in simple cases, one can alleviate this issue by marking the two sticks as separate clusters and making inferences independently; a similar fix is unfeasible for connected, complex geometries. For instance, in a long, thin U-shaped tunnel, two nodes located on the parallel sides of the ``U'' are spatially close but geodesically distant, and hence should not be connected by an edge.
\begin{figure*}[t]
    \centering
    \includegraphics[width = 0.95\linewidth]{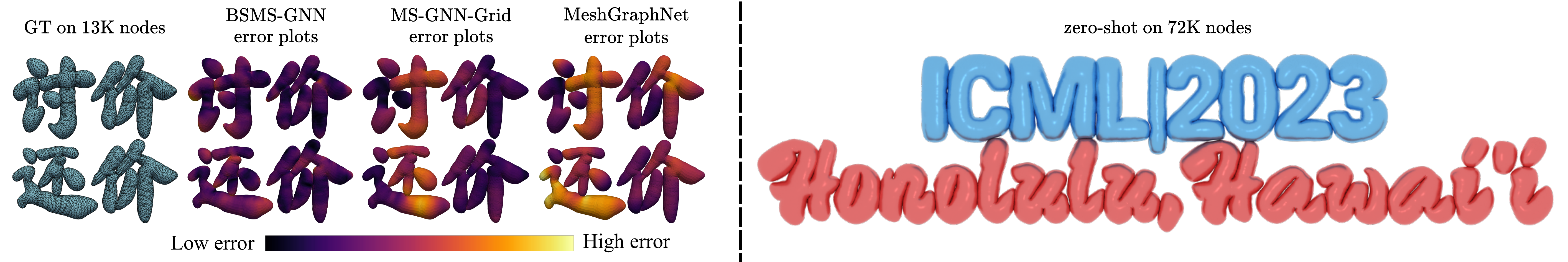}\vspace{-5pt}
    \caption{\textbf{Left:} Comparing the rollout RMSE between \ours~and existing SOTAs on benchmarks \fonts. Our framework reaches the highest accuracy, showing a stronger capability for complex, large geometries with massive contacts. \textbf{Right:} Our trained model can zero-shot infer on never-seen fonts, with about 7x size in mesh size with the same accuracy.}\vspace{-10pt}
    \label{fig:failure:IDP:compares}
\end{figure*}

\paragraph*{Accuracy and Generalization}In terms of accuracy, our method has the smallest rollout RMSE for all cases except for the \plate~dataset, where all three competitors have similar results. \revision{We assume the main reason is that the mesh size for \plate~is too small, so the flat architecture does not show any disadvantage.}
In the largest and most complex dataset, \fonts, our method achieves the highest accuracy, cutting down $40\%$ of the rollout RMSE compared to the competitors (Fig.~\ref{fig:failure:IDP:compares}. Left.). Additionally, our model demonstrates the ability to perform zero-shot inference on larger meshes than those in the training set, and still produce accurate global rollouts, even when the characters are written in a different language.

\paragraph*{Performance Advantages of \ours~}
Our method has a simplified and lightweight model architecture, characterized by a reduced number of MPs at each level and the absence of learnable transition modules. This results in a significant reduction in memory footprint during training (in Fig.\ref{fig:perform:compare}. (c) and Table.\ref{tab:detail:res2}); \ours~consumes $43\% \sim 87\%$ memory in training as \linoGNN, $48\% \sim 53\%$ as \flatGNN, and only $10\%$ as \GUN. our method also uses the least memory during inference, except for the \plate~dataset where the consumption is slightly higher (20MBs) than \flatGNN.

Memory reduction also contributes to improved training efficiency, as it allows for larger batch sizes, more random sampling, and fewer times of data swapping between CPU and GPU. Combined, our method has the fastest unit training speed (per epoch) among all competitors (Fig.\ref{fig:perform:compare}. (a) and Table.\ref{tab:detail:res}), where it consumes only $26\% \sim 58\%$ unit training time as that of \linoGNN~and \flatGNN.

In terms of inference time, our method exhibits similar performance to \linoGNN~on smaller mesh size datasets (\cylinder~and \plate~), both outperforming \flatGNN. However, as the mesh size increases, \ours~surpasses \linoGNN, showing better scalability. In large mesh size cases (\airfoil~and \fonts~), our method shows a $1.5\times$ and $1.9\times$ improvement over \linoGNN~and \flatGNN, respectively (Fig.\ref{fig:perform:compare}. (b) and Table.\ref{tab:detail:res}).

Concerning the total training cost to reach a desired global rollout RMSE, we observe that all methods reach the target with a similar number of epochs. This is because that all these methods learn to resist rollout noise by seeing different, random noises at each epoch; enough noise patterns (proportional to epoch number) is the key for accurate rollout; as such, the total wall time is roughly proportional to the unit training time, given similar epoch numbers, making our method the most efficient.
\begin{figure}[tb]
    \includegraphics[width = 0.9 \linewidth]{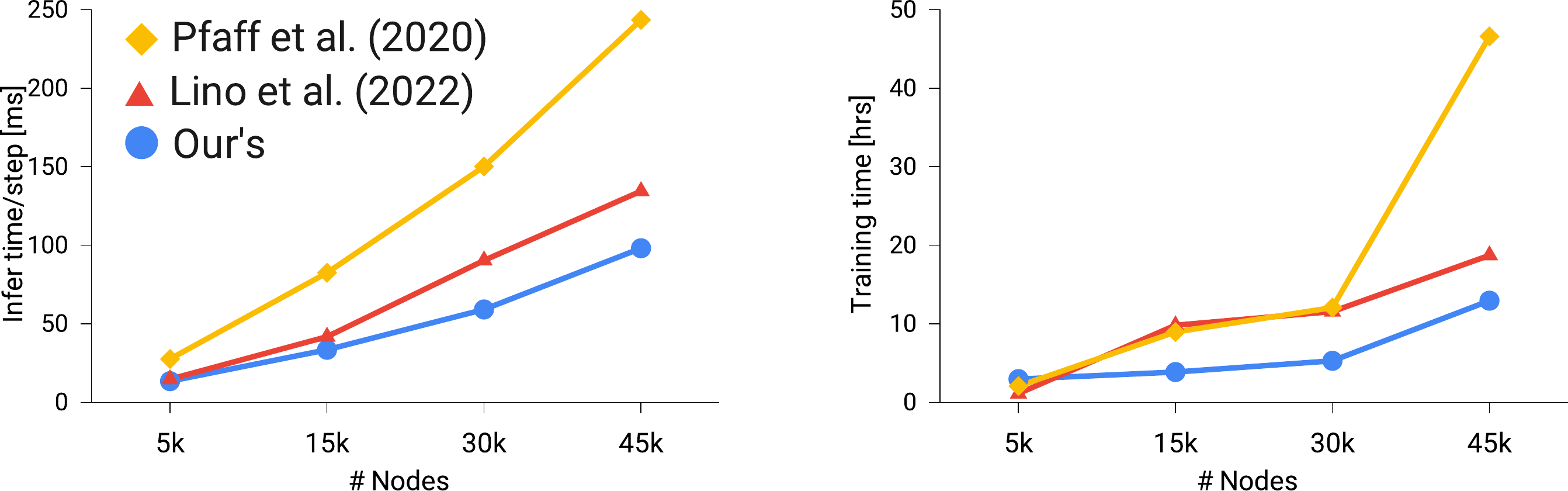}
    \caption{\textbf{Scaling analysis}. With the growing size of \fonts, \ours~shows a progressive advantage over \flatGNN~and \linoGNN.}
    \label{fig:res:scale}
\end{figure}

\paragraph*{Scaling Analysis} By training and evaluating competing methods on \fonts~with varying resolutions (5K,15K,30K, and 45K), we observe that both \ours~and \linoGNN~scale up with a near-linear growing rate; still, our method is more efficient than \linoGNN~as illustrated earlier(Fig.~\ref{fig:res:scale}). We leave the details of scaling analysis in \ref{sec:appdx:scale}.

%% file: s_related_work.tex
\section{Related Works}
\label{sec:related}
\paragraph*{GNNs for Physics-Based Simulation}
The application of GNNs to physics-based simulation has first been applied to deformable solids and fluids (both represented by particles)~\cite{sanchez2018graph}. A notable milestone in this field is \flatGNN~\cite{pfaff2020learning}, which enables the general scheme for learning mesh-based simulations. Subsequently, several variants of \flatGNN~have been proposed: such as combining GNNs with Physics-Informed Neural Networks (PINNs)~\cite{gao2022physics}, making long-term predictions by combining the GraphAutoEncoder (GAE) and Transformer~\cite{han2022predicting}, directly predicting the steady-states through the multi-layer readouts~\cite{harsch2021direct}, and accelerating the finer-level simulation by feeding the up-sampled coarser results inferred by GNN~\cite{belbute2020combining}.

\paragraph*{Multi-Scale GNNs}
Multi-scale GNNs (MS-GNNs) have been widely used in general graph-related tasks other than physics~\cite{wu2020comprehensive,mesquita2020rethinking,zhang2019hierarchical}. The GraphUNet\cite{gao2019graph} introduces the UNet structure into GNNs with a trainable scoring module for pooling, and a $2^{\text{nd}}$-powered adjacency enhancement to help conserve the connectivity. Several works have investigated the use of MS-GNNs for physics-based simulations, including the two- and multi-level GNNs\cite{fortunato2022multiscale,liu2021multi}, which rely on manually drawing coarse meshes. Works such as \linoGNN~\cite{lino2021simulating,lino2022towards} rely on spatial proximity to generate multi-level structures. \citet{li2020multipole} adopts multi-level matrix factorization to generate the kernels at coarser levels. \citet{lino2022multi} utilizes Guillard’s coarsening algorithm to build the coarse-level meshes, but only for 2-D triangle elements. Additionally, there are representative works that abandon the original mesh and build connections and hierarchies on point clouds, such as GNS~\cite{sanchez2020learning}, PointNet~\cite{qi2017pointnet}, PointNet++\cite{qi2017pointnet++}, and GeodesicConv\cite{masci2015geodesic}.

%% file: s_conclusion.tex
\section{Conclusion, Limitations, and Future Work}
The Bi-Stride Multi-Scale Graph Neural Network (BSMS-GNN) utilizes a novel pooling strategy that allows for the creation of an arbitrary-depth, multi-level graph neural network using the original mesh as the sole input. This approach eliminates the need for manually drawing coarser meshes and reduces the potential for wrong edges introduced by spatial proximity. Additionally, Bi-stride pooling enables a one-MP scheme and a non-parametric transition, resulting in a significant reduction in computational costs. Overall, BSMS-GNN improves the capability and generality of applying GNNs to large-scale physical simulations with complex geometries.

\revision{
Further research following our path may include handling huge graphs through the combination of multi-level GNNs with batched and distributed training~\cite{stronisch2023multi}.
Combining the neural operators~\cite{li2020neural,li2020multipole}, i.e. the ability to handle a wide range of PDE parameters, is also appealing.
It would be interesting to combine BSMS-GNNs with Transformer~\cite{han2022predicting,geneva2022transformers,li2022transformer} for stable roll-outs.
Regarding more precise contact modeling, the point-point approach can be improved by face pairs~\cite{allen2023learning}.
Finally, it is worth investigating strategies to score and prunes edges~\cite{ding2006transitive,yu2014transitive} at coarser levels.
}

%% file: s_ack.tex
\subsubsection*{\revision{Acknowledgments}}
\revision{
Yadi Cao would like to thank Na Li for her remote support and companionship throughout his research and study. We thank Neil Shah, Tong Zhao, and Sergey Tulyakov for their invaluable discussions on general GNNs. Yadi Cao would like to acknowledge Dr. Shaowu Pan for the initial inspiration of this work. We extend our appreciation to our reviewers for their valuable feedback on this work and manuscript. This
work has been supported in part by NSF CAREER 2153851, CCF-2153863, ECCS-2023780.
}

%% file: s_appendix.tex
\newpage
\appendix
\onecolumn
\section{Appendix}
\subsection{Dataset details}
\label{sec:appdx:datasets}
We adopt three existing test cases: Cylinder (Flow), Airfoil, and (Deforming) Plate from \flatGNN. The Cylinder includes the transient incompressible flow field around a fixed cylinder at varying locations. The Airfoil includes the transient compressible flow field at varying Mach numbers around the airfoil with varying angles of attack (AOA). The Plate includes hyperelastic plates squeezed by moving obstacles.
In addition to these three cases, our Font(\fonts~) case involves the quasi-static inflation of enclosed elastic surfaces (3D surface mesh) possibly with self-contact. We create the \fonts~cases using the open-source simulator \cite{fang2021guaranteed}, with the same material properties and inflation speed. The input geometries for \fonts~are $1,400$ $2 \times 2$-character matrices in Chinese.
All the datasets are split into 1000 training, 200 validation, and 200 testing instances. In the following table, the second entries with superscript${ }^*$ in the average edge number column are for the contact edges:
\begin{table}[h]
    \centering
    \label{tab:dataset:details}
    \begin{tabular}{ccccccc}
        \textbf{Case} & \textbf{Ave \# nodes} & \textbf{Ave \#  edges} & \textbf{Mesh type} & \textbf{Seed method} & \textbf{\# Levels} & \textbf{\# Steps} \\ \hline
        Cylinder      & 1886                  & 5424                   & triangle, 2D       & MinAve               & 7                  & 600               \\
        Airfoil       & 5233                  & 15449                  & triangle, 2D       & MinAve               & 9                  & 600               \\
        Plate         & 1271                  & $4611, 94^*$           & tetrahedron, 3D    & MinAve               & 6                  & 400               \\
        Font          & 13177                 & $39481, 6716^*$        & triangle, 3D       & CloseCenter          & 6                  & 100
    \end{tabular}
\end{table}

Below we list the model configurations: 1) the offset inputs to prepend before the material edge processor $e^{M}_{ij}$, and $e^{W}_{ij}$, and 2) nodes $p_i$, as well as the nodal outputs $q_i$ from the decoder for each experiment cases, where $\mX$ and $\vx$ stand for the material-space and world-space positions, $\vv$ is the velocity, $\rho$ is the density, $P$ is the absolute pressure, and the dot $\dot{a} = a_{t+1} - a_t$ stands for temporal change for a variable $a$. All the variables involved are normalized to zero-mean and unit variance via pre-processing.
\begin{table}[h]
    \centering
    \label{tab:model:inputs}
    \begin{tabular}{cccccc}
        \textbf{Case} & \textbf{Type} & \textbf{Offset inputs }$e^{M}_{ij}$       & \textbf{Offset inputs }$e^{W}_{ij}$ & \textbf{Inputs }$p_i$ & \textbf{Outputs }$q_i$         \\ \hline
        Cylinder      & Eulerian      & $\mX_{ij},|\mX_{ij}|$                     & NA                                  & $\vv_i,n_i$           & $\dot{\vv}_i$                  \\
        Airfoil       & Eulerian      & $\mX_{ij},|\mX_{ij}|$                     & NA                                  & $\rho_i,\vv_i,n_i$    & $\dot{\vv}_i,\dot{\rho}_i,P_i$ \\
        Plate         & Lagrangian    & $\mX_{ij},|\mX_{ij}|,\vx_{ij},|\vx_{ij}|$ & $\vx_{ij},|\vx_{ij}|$               & $\dot{\vx}_i,n_i$     & $\dot{\vx}_i$                  \\
        Font          & Lagrangian    & $\mX_{ij},|\mX_{ij}|,\vx_{ij},|\vx_{ij}|$ & $\vx_{ij},|\vx_{ij}|$               & $n_i$                 & $\dot{\vx}_i$
    \end{tabular}
\end{table}

As for time integration, Cylinder, Airfoil, and Plate inherited the first-order integration from \flatGNN. For \fonts, the first-order quasi-static integration\ \cite{fang2021guaranteed} is used in the solver. Hence, we also adopt the first-order integration for \fonts.
\subsection{Additional Model details}
\label{sec:appdx:arch}
\subsubsection{Basic modules and architectures}
The MLPs for the nodal encoder, the processor, and the nodal decoder are ReLU-activated two-hidden-layer MLPs with the hidden-layer and output size at 128, except for the nodal decoder whose output size matches the prediction $\vq$. All MLPs have a residual connection. A LayerNorm normalizes all MLP outputs except for the nodal decoder.
\subsubsection{Baseline: \flatGNN~}
Our \flatGNN~implementation uses the same MLPs as above but with an additional module: the edge encoder. Also, the edge latent is updated and carried over throughout the end of multiple MPs. We use 15 times MP for all cases to keep it consistent with \flatGNN.
\subsubsection{Baseline: \linoGNN~}
Our re-implementation of \linoGNN~uses the same MLPs as above but with four additional modules: the edge encoder at the finest level, the aggregation modules for nodes and edges at every level for the transitions, and the returning modules for nodes at every level. This method also requires assigning the regular grid nodes for each level. We assign these grid nodes by defining an initial grid resolution and an inflation rate between levels. As for the MP times at each level, we follow \citet{lino2022towards} to use four at the top and bottom levels and two for the others.
\begin{table}[h]
    \centering
    \label{tab:Lino:detail}
    \begin{tabular}{ccccc}
        \textbf{Case} & \textbf{\# Levels} & \textbf{Initial grid dx}   & \textbf{dx inflation} & \textbf{Level-wise \# MPs} \\ \hline
        Cylinder      & 4                  & {[}5e-2, 5e-2{]}           & 2                     & {[}4, 2, 2, 4{]}           \\
        Airfoil       & 4                  & {[}4.5, 4.5{]}             & 2                     & {[}4, 2, 2, 4{]}           \\
        Plate         & 4                  & {[}4e-3, 4e-3, 4e-3{]}     & 2                     & {[}4, 2, 2, 4{]}           \\
        Font          & 4                  & {[}1.5e-2, 1.5e-2, 1e-3{]} & 2                     & {[}4, 2, 2, 4{]}
    \end{tabular}
\end{table}
\subsubsection{Baseline: \GUN~}
Our re-implementation of \GUN~~uses the same number of levels as those of \ours. Likewise, we make the following modifications to the original \GUN~: (1) We change the information passing from GCN to our message passing module for consistency and translational invariance. (2) GraphUNet was intended for tiny graphs (100 nodes) and used dense matrix multiplications. This design is not scalable as it can break the memory limit and slow down the training to take more than 30 days per epoch in our graph size (1500 to 15000 nodes). We thus optimize the operations such as matrix multiplication and aggregation with sparse implementations.
\subsubsection{Noise and batch number}
\revision{For each of these benchmarks, we generated Gaussian noise with a specific scale and added it to the original trajectory at the beginning of every epoch. The purpose of noise injection was to mimic the effects of noise generated by the model, thereby improving the model's ability to correct its output when fed with noisy inputs. Furthermore, we carefully tuned the batch size under smaller subset experiments for each method to achieve optimal convergence rates.}
\begin{table}[h]
    \centering
    \resizebox{0.8 \textwidth}{!}{
        \centering
        \begin{tabular}{cccccc}
            \textbf{Case} & \multicolumn{4}{c}{\textbf{Batch size}} & \textbf{Noise scale}                                                                      \\
            \textbf{}     & \textbf{\ours~}                          & \textbf{\linoGNN~}    & \textbf{\flatGNN~} & \textbf{\GUN~} & \textbf{}                      \\ \hline
            Cylinder      & 32                                      & 16                   & 16                & 2             & velocity: 2e-2                 \\
            Airfoil       & 8                                       & 4                    & 8                 & 1             & velocity: 2e-2, density: 1e1   \\
            Plate         & 8                                       & 2                    & 2                 & 1             & pos: 3e-3                      \\
            Font          & 2                                       & 1                    & 1                 & 1             & pos: {[}5e-3, 5e-3, 3.33e-4{]}
        \end{tabular}
        \label{tab:noise:batch}
    }
\end{table}

\subsection{Detailed results}
\label{sec:appdx:detail:res}
We plot the detailed measurements in Table.~\ref*{tab:detail:res} and Table.~\ref*{tab:detail:res2}. All experiments are conducted using a single Nvidia RTX 3090.
\begin{table*}[t]
    \centering
    \caption{
        \textbf{Detailed measurements} of our method, \linoGNN, \flatGNN, and \GUN. \ours~consistently generates stable and competitive global rollouts with the smallest training cost. \ours~is also lightweight and has the fastest inference time. It is also free from the large RMSE due to poor pooling on unseen geometries where the learnable pooling module of \GUN~~suffers. The 2nd column of entries in Infer time is the speed up compared to the numerical solver. The 2nd column of entries in Training cost is the epoch for achieving the converged model.}\vspace{5pt}
    \resizebox{0.8\textwidth}{!}{
        \centering
        \begin{tabular}{cccccc}
            \textbf{Measurements}                                                                             & \textbf{Case} & \textbf{Our's}        & \textbf{\cite{lino2021simulating}} & \textbf{\cite{pfaff2020learning}} & \textbf{\cite{gao2019graph}} \\ \hline
            \multirow{4}{*}{\begin{tabular}[c]{@{}c@{}}Training time/step\\ {[}ms{]}\end{tabular}}            & Cylinder      & \textbf{10.14}        & 15.36                              & 19.29                             & 16.20                        \\
                                                                                                              & Airfoil       & \textbf{18.82}        & 25.26                              & 36.72                             & 55.08                        \\
                                                                                                              & Plate         & \textbf{15.58}        & 49.65                              & 49.15                             & 31.88                        \\
                                                                                                              & \fonts~       & \textbf{45.96}        & 107.16                             & 117.48                            & 1,833.37                     \\ \hline
            \multirow{4}{*}{\begin{tabular}[c]{@{}c@{}}Infer time/step\\ {[}ms{]}\end{tabular}}               & Cylinder      & 6.75, 121x                  & \textbf{6.18, 133x}                      & 14.50, 57x                             & 24.30, 34x                        \\
                                                                                                              & Airfoil       & \textbf{8.64, 1275x}         & 20.40, 540x                              & 24.20, 455x                             & 33.60, 328x                        \\
                                                                                                              & Plate         & \textbf{14.01, 207x}        & 18.12, 160x                              & 15.70, 184x                             & 16.20, 179x                        \\
                                                                                                              & \fonts~       & \textbf{33.33, 210x}        & 41.66, 168x                              & 82.35, 85x                             & 629.33, 11x                       \\ \hline
            \multirow{4}{*}{\begin{tabular}[c]{@{}c@{}}Training cost\\ {[}hrs{]},\\ Final epoch\end{tabular}} & Cylinder      & \textbf{21.41, 19}    & 35.84, 21                          & 64.30, 30                         & 76.15, 39                    \\
                                                                                                              & Airfoil       & \textbf{122.33, 39}   & 176.82, 42                         & 275.40, 45                        & 206.55, 37                   \\
                                                                                                              & Plate         & \textbf{56.07, 27}    & 125.78, 19                         & 176.94, 27                        & 127.50, 30                   \\
                                                                                                              & \fonts~       & \textbf{2.68E+01, 21} & 5.66E+01, 19                       & 6.20E+01, 19                      & NA                           \\ \hline
            \multirow{4}{*}{\begin{tabular}[c]{@{}c@{}}RMSE-1\\ {[}1e-2{]}\end{tabular}}                      & Cylinder      & \textbf{2.04E-01}     & 2.20E-01                           & 2.26E-01                          & 8.09E-01                     \\
                                                                                                              & Airfoil       & 2.88E+01              & \textbf{2.68E+01}                  & 4.35E+01                          & 2.93E+01                     \\
                                                                                                              & Plate         & 2.87E-02              & 2.20E-02                           & \textbf{1.98E-02}                 & 2.03E-02                     \\
                                                                                                              & \fonts~       & \textbf{1.77E-02}     & 1.87E-02                           & 1.95E-02                          & NA                           \\ \hline
            \multirow{4}{*}{\begin{tabular}[c]{@{}c@{}}RMSE-50\\ {[}1e-2{]}\end{tabular}}                     & Cylinder      & \textbf{2.42}         & 2.74                               & 4.39                              & 1.87E+01                     \\
                                                                                                              & Airfoil       & \textbf{1.10E+03}     & 1.22E+03                           & 1.66E+03                          & 1.17E+03                     \\
                                                                                                              & Plate         & 3.18E-02              & \textbf{2.78E-02}                  & 2.88E-02                          & 5.19E-02                     \\
                                                                                                              & \fonts~       & \textbf{1.08E-01}     & 3.24E-01                           & 1.78E-01                          & NA                           \\ \hline
            \multirow{4}{*}{\begin{tabular}[c]{@{}c@{}}RMSE-all\\ {[}1e-2{]}\end{tabular}}                    & Cylinder      & \textbf{8.37}         & 8.49                               & 1.07E+01                          & 1.65E+02                     \\
                                                                                                              & Airfoil       & \textbf{4.21E+03}     & 5.56E+03                           & 6.95E+03                          & 6.11E+03                     \\
                                                                                                              & Plate         & 1.60E-01              & \textbf{1.48E-01}                  & 1.51E-01                          & 5.46E-01                     \\
                                                                                                              & \fonts~       & \textbf{2.20E-01}     & 3.78E-01                           & 3.65E-01                          & NA
        \end{tabular}
    }\vspace{-7pt}
    \label{tab:detail:res}
\end{table*}

\begin{table*}[t]
    \centering
    \caption{
        \textbf{Memory footprint under multi-batches}, \ours~consistently cuts RAM consumption by approximately half in all cases in the training stage, and also has the smallest (except for \plate~) inference RAM.}\vspace{5pt}
    \resizebox{0.8\textwidth}{!}{
        \centering
        \begin{tabular}{ccccccccc}
            \textbf{Case}             & \textbf{Method}                    & \multicolumn{6}{c}{\textbf{Training RAM (GBs) with different Batch \#}} & \textbf{Infer RAM (GBs)}                                                                                   \\
                                      & \textbf{}                          & \textbf{2}                                                 & \textbf{4}     & \textbf{8}    & \textbf{16}    & \textbf{32}   & \textbf{64}    &               \\ \hline
            \multirow{4}{*}{\cylinder} & \textbf{Our's}                     & \textbf{2.41}                                              & \textbf{2.92}  & \textbf{4.37} & \textbf{6.06}  & \textbf{11.4} & \textbf{22.27} & \textbf{1.92} \\
                                      & \textbf{\cite{lino2021simulating}} & 2.79                                                       & 3.60           & 5.31          & 8.56           & 15.10         & -              & 1.97          \\
                                      & \textbf{\cite{pfaff2020learning}}  & 3.25                                                       & 4.46           & 6.91          & 11.84          & 21.60         & -              & 1.94          \\
                                      & \textbf{\cite{gao2019graph}}       & 23.33                                                      & -              & -             & -              & -             & -              & 2.18          \\ \hline
            \multirow{4}{*}{\airfoil}  & \textbf{Our's}                     & \textbf{3.66}                                              & \textbf{5.46}  & \textbf{8.88} & \textbf{15.70} & -             & -              & \textbf{2.02} \\
                                      & \textbf{\cite{lino2021simulating}} & 4.18                                                       & 6.25           & 10.65         & 19.25          & -             & -              & \textbf{2.02} \\
                                      & \textbf{\cite{pfaff2020learning}}  & 5.53                                                       & 8.90           & 16.08         & -              & -             & -              & 2.06          \\
                                      & \textbf{\cite{gao2019graph}}       & -                                                          & -              & -             & -              & -             & -              & 2.67          \\ \hline
            \multirow{4}{*}{\plate}    & \textbf{Our's}                     & \textbf{2.36}                                              & \textbf{2.87}  & \textbf{3.85} & \textbf{5.78}  & \textbf{9.28} & \textbf{16.85} & 1.95          \\
                                      & \textbf{\cite{lino2021simulating}} & 3.41                                                       & 4.81           & 7.75          & 13.20          & -             & -              & 2.00          \\
                                      & \textbf{\cite{pfaff2020learning}}  & 3.10                                                       & 4.29           & 6.59          & 11.49          & 20.80         & -              & \textbf{1.93} \\
                                      & \textbf{\cite{gao2019graph}}       & -                                                          & -              & -             & -              & -             & -              & 2.18          \\ \hline
            \multirow{4}{*}{\fonts}     & \textbf{Our's}                     & \textbf{6.28}                                              & \textbf{10.80} & -             & -              & -             & -              & \textbf{2.23} \\
                                      & \textbf{\cite{lino2021simulating}} & 10.87                                                      & 19.79          & -             & -              & -             & -              & 2.45          \\
                                      & \textbf{\cite{pfaff2020learning}}  & 12.48                                                      & 23.39          & -             & -              & -             & -              & 2.28          \\
                                      & \textbf{\cite{gao2019graph}}       & -                                                          & -              & -             & -              & -             & -              & 4.51
        \end{tabular}\vspace{-7pt}
    }
    \label{tab:detail:res2}
\end{table*}
\subsection{Ablation study}
\label{sec:appdx:abaltion}
\subsubsection{Transition method}
\label{sec:appdx:abaltion:trans}
\begin{figure}[t]
    \includegraphics[width = \linewidth]{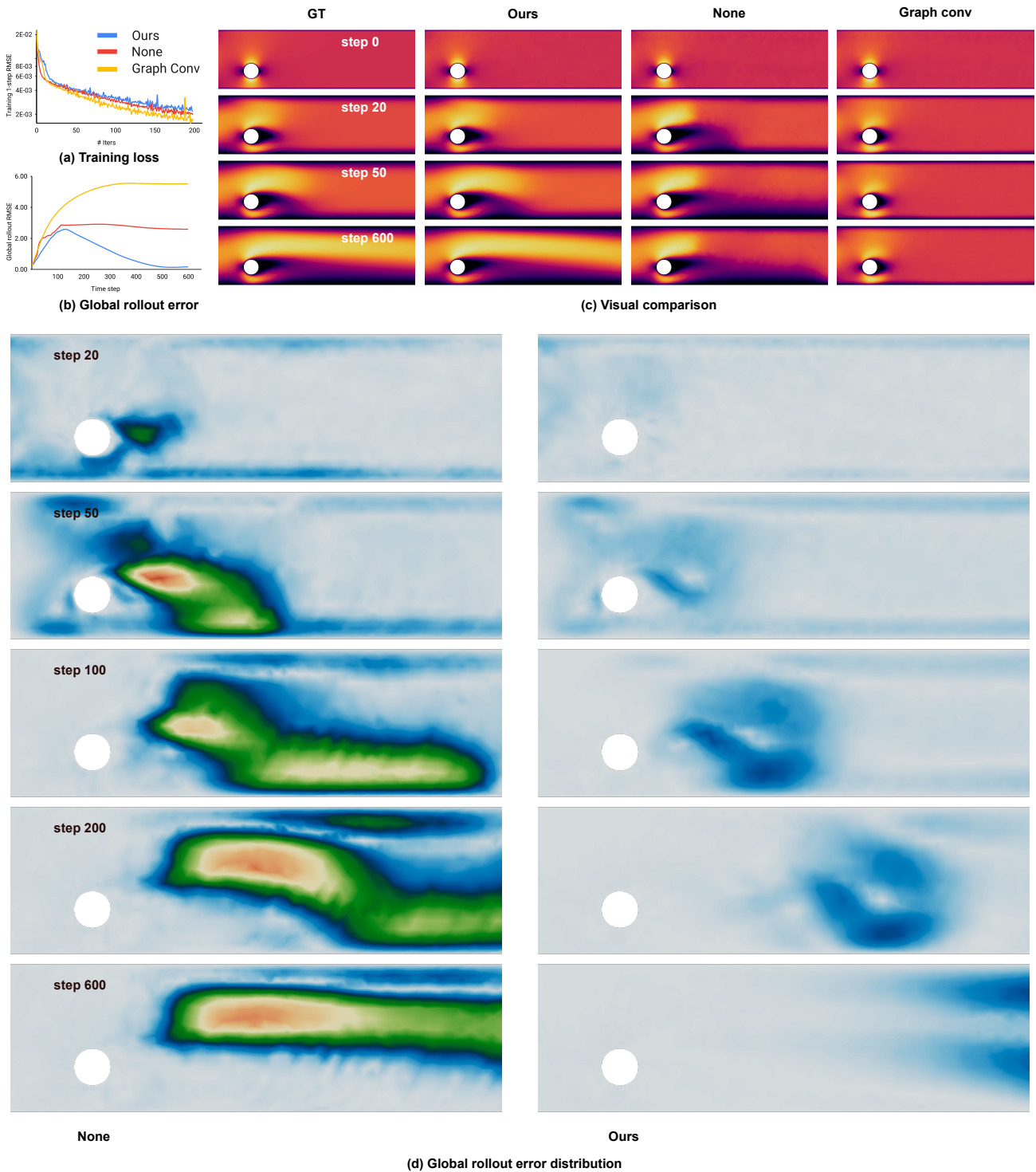}
    \caption{(a) All three transition methods can reach the target training RMSE given 200 iterations. (b) However, our weighted graph aggregration+returning has the strongest resistance to the noise during the rollout. (c) The visual comparisons show that no transition produces mosaic-like patterns, while the graph convolution transition smeared out the information and ceased propagating downstream. (d) The global rollout error distribution of no transition (\textbf{Left}) shows the edge of the mosaic patterns look similar to the simulation mesh; The error of our transition (\textbf{Right}) travels with the generated vortices downstream and leaves the domain after step 200, which explains the RMSE drop in (b).}
    \label{fig:ablation:trans}
\end{figure}
While exploring the non-parametric transition solutions, we started with no transition because our method is adopted directly from GUN \cite{gao2019graph}. The no-transition strategy produces low enough 1-step RMSE and visually correct rollouts for \fonts. However, in the global rollouts of \cylinder~and \airfoil~cases, we observed stripe patterns (Figure.~\ref{fig:ablation:trans} (c), column \textbf{None}) where the stripes are aligned with the edges at the coarser levels (Figure.~\ref{fig:ablation:trans} (d)). We suspect that this error results from the fact that the unpooled nodes all have zero information before MP, making them indistinguishable from the processor modules and exaggerating the difference between pooled and unpooled nodes over rollouts.

The no-transition strategy resembles no interpolation during the super-resolution phase of CNN+UNet. Naturally, we then tried a single step of graph convolution (without activation) to resemble the interpolation in regular grids. However, this turns out to over-smooth the features (Figure.~\ref{fig:ablation:trans} (e), column \textbf{Graph Conv}), and the information propagation was smeared out except for the area near the generator (in this case, near the cylinder).

We believe the over-smoothing issue arises from the ignorance of the irregularity of the mesh. Unlike CNN, where the fine nodes regularly lie at the center of coarser grids, irregular meshes have varying topology and element sizes. The element sizes are almost always smaller near the interface for higher precision in simulations; hence an unweighted graph convolution can smear the finer information near the cylinder and their adjacent neighbors during returning. The natural choice to account for the irregularity is to include reasonable nodal weights (such as the size). In the end, we arrive at the solution proposed in Sec.~\ref{sec:transition} by utilizing the nodal weights during aggregation and recording the shares of contribution for later returning. Our transition method works consistently for all experiment cases and produces the lowest RMSE for global rollouts (Figure.~\ref{fig:ablation:trans} (b)).

\paragraph*{Comparing to alternative transition methods} Additionally, we compare our transition methods to two alternatives extracted from previous works: (1) calculating the edge weights (kernel) for the information flow using the inverse of its length (node position offset), which we refer to as \textbf{Pos-Kernel} \cite{liu2021multi}; and (2) the level-wise learnable transition modules implemented by additional MP, which we refer to as \textbf{Learnable} \cite{fortunato2022multiscale}.

\begin{table}[h]
    \centering
    \caption{\textbf{Detailed measurements} of different transition methods. Ours and \textbf{Pos-Kernel} are the only two non-parametric transitions that are light-weighted and produce reliable rollouts compared to the expensive \textbf{Learnable} transition.}\vspace{5pt}
    \begin{tabular}{lccccc}
        \textbf{Measurements}       & \textbf{Ours}   & \textbf{None}     & \textbf{Graph-Conv} & \textbf{Pos-Kernel} & \textbf{Learnable} \\ \hline
        Training time/step {[}ms{]} & 10.14           & \textbf{9.30}     & 10.07               & 10.06               & 17.75              \\
        Infer time/step {[}ms{]}    & 6.75            & \textbf{5.70}     & 6.46                & 6.90                & 11.28              \\
        Training RAM {[}GBs{]}      & \textbf{11.041} & \textbf{11.041}   & \textbf{11.041}     & \textbf{11.041}     & 18.033             \\
        Infer RAM {[}GBs{]}         & \textbf{1.923}  & \textbf{1.923}    & \textbf{1.923}      & \textbf{1.923}      & 1.931              \\
        RMSE-1 {[}1e-2{]}           & 2.85E-01        & \textbf{1.49E-01} & 3.41E-01            & 6.38E-01            & 4.70E-01           \\
        RMSE-50 {[}1e-2{]}          & 1.43E+01        & 2.05E+02          & 2.40E+02            & 1.77E+01            & \textbf{1.35E+01}  \\
        RMSE-all {[}1e-2{]}         & 1.68E+01        & 2.59E+02          & 5.51E+02            & 2.01E+01            & \textbf{1.57E+01}
    \end{tabular}
    \label{tab:detail:ram}
\end{table}

In addition to the high RMSE of \textbf{None} and \textbf{Graph-Conv} shown in Figure.~\ref{fig:ablation:trans}, we can also observe that: (1) the training/infer time and RAM consumption for all non-parametric transitions (including \textbf{None}) are similar, which supports the statement that our transition method is light-weighted. (2) \textbf{Learnable} transition can reach slightly higher accuracy but at the price of $\sim 70\%$ more training/infer time and RAM. As mentioned in Sec.~\ref{sec:experiments:results}, higher training RAM can limit the batch number and increase the frequency of data communication between CPU and GPU, slowing down the training process even further when the scale goes up. (3) \textbf{Pos-Kernel} results in a slightly higher RMSE compared to our method, making it a competitive alternative in production.

\revision{\subsubsection{Sensitivity analysis on seeding heuristics}}
\label{sec:appdx:abaltion:seeding}
\revision{
In this section, we investigated the impact of using different seeding heuristics during the training and testing phases on the sensitivity of a converged model. We deliberately altered the heuristics used on each benchmark and evaluated the RMSEs. The results showed that the inconsistency in seeding heuristics led to higher roll-outs compared to the results obtained when using consistent seeding, as shown in Table~\ref{tab:ablation:seeding}. Specifically, the roll-outs were 1.01x to 2.04x and 1.13x to 2.21x for random seeding and inverse seeding, respectively, with respect to consistent seeding. However, the RMSEs remained in the same magnitude, indicating that our method is not sensitive to the initial seeding.
It should be also noted that this is not a practical issue, as the seeding can easily be kept consistent during different phases.
}

\begin{table}[h]
    \centering
    \centering
    \caption{\revision{\textbf{Sensitivity analysis of seeding heuristics on model performance.} The ``Random'' heuristic refers to choosing a random seed for every cluster, while ``Inverse w.r.t. Train'' refers to choosing the inverse seeding heuristic compared to that used during the training phase. For example, if the MinAve heuristic was used during training, the CloseCenter heuristic was chosen during the testing phase.}}
    \label{tab:ablation:seeding}
\begin{tabular}{c|ccccc}
\textbf{Seeding @ Test}                        & \textbf{Ratio in RMSE} & \textbf{Cylinder} & \textbf{Airfoil} & \textbf{Plate} & \textbf{Font} \\ \hline
\multirow{3}{*}{\textbf{Random}}               & 1-step                 & 2.06              & 14.04            & 2.95           & 1.15          \\
                                               & 50-step                & 1.06              & 2.16             & 5.21           & 1.04          \\
                                               & Rollout                & 1.01              & 2.04             & 1.53           & 1.04          \\ \hline
\multirow{3}{*}{\textbf{Inverse w.r.t. Train}} & 1-step                 & 1.96              & 12.77            & 3.14           & 1.15          \\
                                               & 50-step                & 1.07              & 1.76             & 7.52           & 1.02          \\
                                               & Rollout                & 1.13              & 2.21             & 1.46           & 1.06         
\end{tabular}
\end{table}

\subsection{Details for scaling analysis on \fonts}
\label{sec:appdx:scale}
We generate the downscale and the upscale version of \fonts~with different average node numbers for the initial geometry, and then use the same settings to simulate the sequence. As reported in \citet{fortunato2022multiscale}, the low-resolution model suffers from converging to very small RMSE; hence we loosen the termination criteria by enlarging the target RMSE relative to the average edge length to prevent convergence failures. Similarly, the noise injection is also adjusted to be relative to the average edge length. Moreover, with a smaller number of nodes, the number of levels required to achieve the same bottom resolution also reduces. We make the corresponding adjustments to the levels of our model $d_1$ and that of the \linoGNN~$d_2$. The adjustments are plotted below.

\begin{table}[h]
    \centering
    \caption{\textbf{The adjustment for multi-scale parameters, the target RMSE and noise injection for scaling analysis.}}
    \label{tab:scale:model:detail}
    \begin{tabular}{cccccc}
        \textbf{\# Nodes} & $d_1$ & $d_2$ & \textbf{Initial grid dx} & \textbf{Target RMSE} & \textbf{Noise in pos}        \\ \hline
        5k                & 4     & 2     & {[}6e-2 6e-2 4e-3{]}     & 1.73e-4              & {[}8.5e-3, 8.5e-3, 5.7e-4{]} \\
        15k               & 6     & 4     & {[}1.5e-2 1.5e-2 1e-3{]} & 1e-4                 & {[}5e-3, 5e-3, 3.33e-4{]}    \\
        30k               & 7     & 5     & {[}7.5e-3 7.5e-3 5e-4{]} & 1e-4                 & {[}3.5e-3, 3.5e-3, 2.4e-4{]} \\
        45k               & 7     & 5     & {[}7.5e-3 7.5e-3 5e-4{]} & 1e-4                 & {[}2.9e-3, 2.9e-3, 1.9e-4{]}
    \end{tabular}
\end{table}

\subsection{The Proof of conservation of contact edges}
\label{sec:appdx:prove:contact:conserve}
With Bi-stride pooling, our pooling conserves all the contact edges under the enhancement in Eq.~\ref{eq:adj:enhance}. We assume the graph is undirected and unweighted, such that the adjacent matrix is a boolean matrix.

Formally speaking, given any contact edge $(i,j)$ at level $l$ (i.e. $\mA^C_l[i,j]=1$) and a Bi-stride pooling $P$ which pools nodes $\mathcal{I}$, there exists a contact edge $(i',j')$ that remains in the coarser level (i.e. ${\mA'}^C_{l+1}[i',j']=1, i',j' \in \mathcal{I}$) and $i/i',j/j'$ are connected (i.e. $\mA_l[i,i']=\mA_l[j,j']=1$). There are only four scenarios concerning the pooling nodes $\mathcal{I}$ and the contact edge nodes $i,j$, under which the assertion always holds:
\begin{enumerate}
    \item Both $i,j$ are pooled, i.e. $i,j \in \mathcal{I}$. Obviously ${\mA'}^C_{l+1}[i',j']=1$ by letting $i'=i, j'=j$.
    \item Only $i$ is pooled, $i \in \mathcal{I}, j\notin \mathcal{I}$. Since we use Bi-stride pooling, $j$ can either be the seed at level 0 (Bi-stride can select either even or odd levels) that directly connects to all nodes at level 1, or must have at least one direct connection from the previous level. I.e, at least one neighbor of $j$ in the adjacent level is pooled, we let it be $j'$: $\mA_l[j,j']=1, j' \in \mathcal{I}$. Then $\mA^C_l \mA_l [i,j'] \geq \mA^C_l[i,j] * \mA_l[j,j'] = 1$, and $\mA_l (\mA^C_l \mA_l) [i,j'] \geq \mA_l[i,i] * (\mA^C_l \mA_l) [i,j']=1$. Let $i'=i$, then ${\mA'}^C_{l+1}[i',j']=1$.
    \item Only $j$ is pooled, $i \notin \mathcal{I}, j \in \mathcal{I}$. Similarly we have at least one $i'$ such that: $\mA_l[i',i]=1, i' \in \mathcal{I}$. Then $\mA_l \mA^C_l [i',j] \geq \mA_l[i',i] * \mA^C_l[i,j] = 1$, and $(\mA_l \mA^C_l) \mA_l [i',j] \geq (\mA_l \mA^C_l)[i',j] * \mA_l [j,j]=1$. Let $j'=j$, then ${\mA'}^C_{l+1}[i',j']=1$.
    \item None of $i,j$ is pooled, $i,j \notin \mathcal{I}$. We select one direct pooled neighbor for $i,j$, respectively, that $\mA_l[i',i]=\mA_l[j,j']=1, i',j' \in \mathcal{I}$. Then $\mA_l \mA^C_l [i',j] \geq \mA_l[i',i] * \mA^C_l[i,j] = 1$, and $(\mA_l \mA^C_l) \mA_l [i',j'] \geq (\mA_l \mA^C_l)[i',j] * \mA_l [j,j']=1$.
\end{enumerate}

\subsection{Algorithms for the seeding heuristics}
\label{sec:appdx:algo}
Here we elaborate our two seeding heuristics for the bi-stride pooling at every level: picking the seed that 1) is closest to the center of a cluster (CloseCenter), and 2) with the minimum average geodesic distance to its neighbors (MinAve). The complexity for MinAve is $\mO(N^2)$ as we need to conduct BFS for every node to find the one with the minimum average distance to neighbors. In our experiments, the quadratic cost of MinAve is tolerable for all cases but \fonts.

\begin{algorithm}[h]
    \begin{algorithmic}
        \STATE {\bfseries Input:}Unweighted, Bi-directional graph, $\mG=(N,E)$
        \STATE \COMMENT{List of seeds in each clusters $L_s$}
        \STATE $L_c \leftarrow \text{DetermineCluster}(\mG) $
        \STATE $L_s \leftarrow \emptyset $
        \STATE \COMMENT{BFS$(s)$ returns the list of distances to all other neighbors from $s$}
        \STATE \COMMENT{if unreachable, the distance is set to infinity}
        \STATE $D \leftarrow \{ \text{BFS}(s) \text{ for } s \text{ in } N\} $
        \FOR{$\text{idx in } L_c$}
        \STATE $D_c \leftarrow D[\text{idx},\text{idx}]$
        \STATE $\bar{D}_c \leftarrow \text{average}(D_c,\text{dim}=1)$
        \STATE $s \leftarrow \text{idx}[\text{argmin}(\bar{D}_c)]$
        \STATE $L_s.\text{append}(s)$
        \ENDFOR
        \OUTPUT{$L_s$}
    \end{algorithmic}
    \caption{MinAve: seeding by minimum average geodesic distance to neighbors}
\end{algorithm}

For \fonts, the largest mesh has around 47K nodes, and the time for pre-processing with MinAve becomes intolerable. We switch to CloseCenter with the linear complexity.

\begin{algorithm}[h]
    \begin{algorithmic}
        \STATE {\bfseries Input:} Unweighted, Bi-directional graph, $\mG=(N,E)$; Positions of the nodes, $X$
        \STATE \COMMENT{List of seeds in each clusters $L_s$}
        \STATE $L_c \leftarrow \text{DetermineCluster}(\mG) $
        \STATE $L_s \leftarrow \emptyset $
        \FOR{idx {\bfseries in} $L_c$}
        \STATE $\bar{X} \leftarrow \text{average}(X[\text{idx}],\text{dim}=0)$
        \STATE $\Delta X \leftarrow X - \bar{X}$
        \STATE $D \leftarrow ||\Delta X||_2$
        \STATE $s \leftarrow \text{idx}[\text{argmin}(D)]$
        \STATE $L_s.\text{append}(s)$
        \ENDFOR
        \OUTPUT{$L_s$}
    \end{algorithmic}
    \caption{CloseCenter: seeding by minimum distance to the center of cluster}
\end{algorithm}

For both heuristics, we search seeds in a per-cluster fashion to avoid the information from other clusters that could pollute the search result. For example, when determining the center of an isolated part of the input geometry, the positions of nodes from other clusters could pollute this process. The determination of clusters given in a graph is elaborated below.

\begin{algorithm}[h]
    \begin{algorithmic}
        \STATE {\bfseries Input:} Unweighted, Bi-directional graph, $\mG=(N,E)$
        \STATE \COMMENT{$R$ stands for remaining nodes that are not inside any cluster}
        \STATE $R \leftarrow N $
        \STATE $L_c \leftarrow \emptyset $
        \WHILE{$R \neq \emptyset$}
        \STATE $s \leftarrow R.\text{pop}(~)$
        \IF{$|R| = 0$}
        \STATE $L_C.\text{append}(\{c\})$
        \ELSE
        \STATE $D \leftarrow \text{BFS}(s)$
        \STATE $C \leftarrow \emptyset $
        \STATE $R^* \leftarrow \emptyset$
        \FOR{$n$ in $R$}
        \IF{$D[n] = \infty$}
        \STATE $R^*.\text{append}(n)$
        \ELSE
        \STATE $C.\text{append}(n)$
        \ENDIF
        \ENDFOR
        \STATE $L_C.\text{append}(C)$
        \STATE $R \leftarrow R^*$
        \ENDIF
        \ENDWHILE
        \OUTPUT $L_c$
    \end{algorithmic}
    \caption{DetermineCluster}
\end{algorithm}